\documentclass[11pt,a4paper]{article}
\usepackage[hyperref]{acl2021}
\usepackage{times}
\usepackage{latexsym}

\aclfinalcopy

\usepackage{microtype}
\usepackage{comment} 
\usepackage{enumitem}
\usepackage{graphicx}
\usepackage{url}
\usepackage{amsmath}
\usepackage{float}
\usepackage{xcolor}
\usepackage{subcaption}
\usepackage{multirow}
\usepackage{amsfonts}
\usepackage{booktabs}
\usepackage{bbm}
\usepackage{makecell}
\usepackage{pifont}
\usepackage{makecell,rotating}
    \setcellgapes{5pt}

\def\aclpaperid{74} 

\hypersetup{colorlinks,urlcolor={blue}}

\newcommand{\footnoteref}[1]{\textsuperscript{\ref{#1}}}

\makeatletter
\def\thickhline{%
  \noalign{\ifnum0=`}\fi\hrule \@height \thickarrayrulewidth \futurelet
   \reserved@a\@xthickhline}
\def\@xthickhline{\ifx\reserved@a\thickhline
               \vskip\doublerulesep
               \vskip-\thickarrayrulewidth
             \fi
      \ifnum0=`{\fi}}
\makeatother

\newlength{\thickarrayrulewidth}
\title{SAPPHIRE: Approaches for Enhanced Concept-to-Text Generation}

\author{Steven Y. Feng, Jessica Huynh, Chaitanya Narisetty, Eduard Hovy, Varun Gangal\\
        Language Technologies Institute \\ Carnegie Mellon University \\ {\tt \{syfeng,jhuynh,cnariset,hovy,vgangal\}@cs.cmu.edu}}

\date{}

\begin{document}
\maketitle

\begin{abstract}
We motivate and propose a suite of simple but effective improvements for concept-to-text generation called SAPPHIRE: Set Augmentation and Post-hoc PHrase Infilling and REcombination. 
We demonstrate their effectiveness on generative commonsense reasoning, a.k.a. the \emph{CommonGen} task, through experiments using both BART and T5 models. Through extensive automatic and human evaluation, we show that SAPPHIRE noticeably improves model performance. An in-depth qualitative analysis illustrates that SAPPHIRE effectively addresses many issues of the baseline model generations, including lack of commonsense, insufficient specificity, and poor fluency.
\end{abstract}

\section{Introduction}
\label{sec:intro}
There has been increasing interest in constrained text generation tasks which involve constructing natural language outputs under certain pre-conditions, such as particular words that must appear in the output sentences. A related area of work is data-to-text natural language generation (NLG), which requires generating natural language descriptions of structured or semi-structured data inputs. Many constrained text generation and NLG tasks share commonalities, one of which is their task formulation: a set of inputs must be converted into natural language sentences. This set of inputs can be, in many cases, thought of as \textit{concepts}, e.g. higher-level words or structures that play an important role in the generated text.

With the increased popularity of Transformer-based models and their application to many NLP tasks, performance on many text generation tasks has improved considerably. Much progress in recent years has been from the investigation of model improvements, such as larger and more effectively pretrained language generation models. However, are there simple and effective approaches to improving performance on these tasks that can come from the data itself? Further, can we potentially use the outputs of these models themselves to further improve their task performance - a ``self-introspection" of sorts?

In this paper, we show that the answer is yes. We propose a suite of simple but effective improvements for concept-to-text generation called SAPPHIRE: Set Augmentation and Post-hoc PHrase Infilling and REcombination. Specifically, SAPPHIRE is composed of two major approaches: 1) the augmentation of input concept sets (\S\ref{sec:concept_set_aug}), 2) the recombination of phrases extracted from baseline generations into more fluent and logical text (\S\ref{sec:phrase_recombination}). These are mainly model-agnostic improvements that rely on the data itself and the model's own initial generations, respectively.\footnote{Code at \url{https://github.com/styfeng/SAPPHIRE}}

We focus on generative commonsense reasoning, or CommonGen \cite{lin-etal-2020-commongen}, which involves generating logical sentences describing an everyday scenario from a set of concepts, which in this case are individual words that must be represented in the output text in some form. CommonGen is a challenging instance of constrained text generation that assesses 1) relational reasoning abilities using commonsense knowledge, and 2) compositional generalization capabilities to piece together concept combinations. Further, CommonGen's task formulation and evaluation methodology are quite broadly applicable and encompassing, making it a good benchmark for general constrained text generation capability. Further, this is an opportune moment to investigate this task as commonsense ability of NLP models, particularly for generation, has received increasing community attention through works like COMET \cite{bosselut-etal-2019-comet}.

We perform experiments on varying sizes of two state-of-the-art Transformer-based language generation models: BART \cite{lewis-etal-2020-bart} and T5 \cite{JMLR:v21:20-074}. We first conduct an extensive correlation study (\S\ref{sec:correlation_study}) and qualitative analysis (\S\ref{sec:initial_qualitative_analysis}) of these models' generations after simply training on CommonGen. We find that performance is positively correlated with concept set size, motivating concept set augmentation. We also find that generations contain issues related to commonsense and fluency which can possibly be addressed through piecing the texts back together in different ways, motivating phrase recombination.



Fleshing out our first intuition - we devise two methods to augment concepts from references during training through extracted keywords (\S\ref{sec:kw-aug}) and attention matrices (\S\ref{sec:att-aug}). For the phrase recombination intuition, we propose two realizations based on a new training stage (\S\ref{sec:P2T}) and masked infilling (\S\ref{sec:mask_infilling}). Finally, through comprehensive evaluation (\S\ref{sec:results_and_analysis}), we show how the SAPPHIRE suite drives up model performance across metrics, besides addressing aforementioned baseline deficiencies on commonsense, specificity, and fluency.



\section{Dataset, Models, and Metrics}
\label{sec:dataset_models_metrics}
\subsection{CommonGen Dataset}
The CommonGen dataset is split into train, dev, and test splits, covering a total of 35,141 concept sets and 79,051 sentences. The concept sets range from 3 keywords to 5 keywords long. As the original test set is hidden, 
we split the provided dev set into a new dev and test split for the majority of our experiments while keeping the training split untouched. Note that we also evaluate our SAPPHIRE models on the original test set with help from the CommonGen authors (see \S\ref{sec:automatic_eval}). We will henceforth refer to these new splits as train$_{CG}$, dev$_{CG}$, and test$_{CG}$, and the original dev and test splits as dev$_{O}$ and test$_{O}$. The statistics of our new splits compared to the originals can be found in Table \ref{tab:dataset_stats}. We attempt to keep the relative sizes of the new dev and test splits and the distribution of concept set sizes within each split similar to the originals.

\begin{table}[t]
\centering
\small
\resizebox{\columnwidth}{!}{\begin{tabular}{ |p{1.8cm}|p{0.9cm}|p{0.6cm}p{0.6cm}|p{0.65cm}p{0.7cm}| }
\hline
\textbf{Dataset Stats} & Train$_{CG}$ & Dev$_{O}$ & Test$_{O}$ & Dev$_{CG}$ & Test$_{CG}$ \\
\hline
\# concept sets & 32,651 & 993 & 1,497 & 240 & 360\\
  size = 3 & 25,020 & 493 & - & 120 & -\\
  size = 4 & 4,240 & 250 & 747 & 60 & 180\\
  size = 5 & 3,391 & 250 & 750 & 60 & 180\\
\hline
\# sentences & 67,389 & 4,018 & 7,644 & 984 & 1583\\
\hline
\end{tabular}}
\vspace{-2mm}
\caption{\footnotesize CommonGen dataset statistics.
}
\label{tab:dataset_stats}
\end{table}



\subsection{Models: T5 and BART}\label{sec:models_T5_BART}
We perform experiments using pretrained language generators, specifically BART and T5 (both base and large versions). BART \cite{lewis-etal-2020-bart} is a Transformer-based seq2seq model trained as a denoising autoencoder to reconstruct original text from noised text. T5 \cite{JMLR:v21:20-074} is another seq2seq Transformer with strong multitask pretraining. We use their HuggingFace codebases. 



We train two seeded instances of each model on train$_{CG}$, evaluating their performance on dev$_{O}$, and comparing our numbers to those reported in \citet{lin-etal-2020-commongen} to benchmark our implementations. These essentially serve as the four baseline models for our ensuing experiments. We follow the hyperparameters from \citet{lin-etal-2020-commongen}, choose the epoch reaching highest ROUGE-2 on the dev split, and use beam search for decoding.\footnote{See Appendix \ref{appendix:model_training_finetuning_details} for further details.} From Table \ref{tab:reimplementation_stats}, we see that our re-implemented models match or exceed the original reported results on most metrics across different models.

\begin{table}[t]
\centering
\small
\begin{tabular}{|c|c|c|c|}
\hline
 \textbf{Model\textbackslash Metrics} &  \multicolumn{1}{c|}{BLEU-4} & CIDEr  & SPICE  \\
 \hline
 Reported BART-large    & 27.50 & 14.12  & 30.00   \\
 \hline
 Reported T5-base    & 18.00 & 9.73  & 23.40   \\
 \hline
 Reported T5-Large    & 30.60 & 15.84  & 31.80   \\
 \thickhline
 Our BART-base    & 28.30 & 15.07  & 30.35   \\
 \hline
 Our BART-large    & \textbf{30.20} & \textbf{15.72}  & \textbf{31.20}   \\ 
 \hline
 Our T5-base    & \textbf{31.00} & \textbf{16.37}  & \textbf{32.05}   \\
 \hline
 Our T5-large    & \textbf{33.60} & \textbf{17.02}  & \textbf{33.45}   \\
 \hline
\end{tabular}
\vspace{-2mm}
\caption{ \footnotesize Performance of our re-implemented CommonGen models on dev$_{O}$ compared to a subset of original numbers reported in \citet{lin-etal-2020-commongen}. For our models, results are averaged over two seeds. The original authors did not experiment with BART-base. Bold indicates where we match or exceed the reported metric. See \S\ref{sec:eval_metrics} for explanations of the metrics and Appendix \ref{appendix:reimplementation_numbers} for a full metric comparison table.}
\label{tab:reimplementation_stats}
\end{table}


\subsection{Evaluation Metrics}\label{sec:eval_metrics}
For our experiments, we use a gamut of automatic evaluation metrics. These include those used by \citet{lin-etal-2020-commongen}, such as 
BLEU \cite{papineni2002bleu}, 
CIDEr \cite{vedantam2015cider}, SPICE \cite{anderson2016spice}, and Coverage (Cov). Barring Cov, these metrics measure the similarity between generated text and human references. Cov measures the average \% of input concepts covered by the generated text. 
We also introduce BERTScore \cite{zhang2019bertscore}, which measures token-by-token BERT \cite{devlin-etal-2019-bert} embeddings similarity. It also measures the similarity between the generated text and human references, but on a more semantic (rather than surface token) level. When reporting BERTScore, we multiply by 100. For all metrics, higher corresponds to better performance.

\section{Initial Analysis}
\label{sec:initial_analysis}
\subsection{Correlation Study}\label{sec:correlation_study}
We begin by conducting an analysis of the four baselines implemented and discussed in \S\ref{sec:models_T5_BART}, which we refer to henceforth as BART-base-BL, BART-large-BL, T5-base-BL, and T5-large-BL. One aspect we were interested in is whether the number of input concepts affects the quality of generated text. We conduct a comprehensive correlation study of the performance of the four baselines on dev$_{O}$ w.r.t. the number of input concepts. 

As seen from Table \ref{tab:correlations}, the majority of the metrics are positively correlated with concept set size across the models. ROUGE-L, CIDEr, and SPICE have small correlations that are mainly statistically insignificant, demonstrating that they are likely uncorrelated with concept set size. Coverage is strongly negatively correlated, showing that there is a higher probability of concepts missing from the generated text as concept set size increases.

There are two major takeaways from this. Firstly, increased concept set size results in greater overall performance. Secondly, models have difficulty with coverage given increased concept set size. This motivates our first set of improvements, which involves augmenting the concept sets with additional words in hopes of 1) increasing performance of the models and 2) improving their coverage, as we hope that training with more input concepts will help models learn to better cover them in the generated text. This is discussed more in \S\ref{sec:concept_set_aug}.


\begin{table*}[t]
\centering
\small
\resizebox{\textwidth}{!}{\begin{tabular}{ |p{1.4cm}|c|c|c|c|c|c|c|c|c|c|c|c| }
\hline
 & \multicolumn{3}{c|}{\textbf{BART-base}} & \multicolumn{3}{c|}{\textbf{BART-large}} & \multicolumn{3}{c|}{\textbf{T5-base}} & \multicolumn{3}{c|}{\textbf{T5-large}}\\
 \hline
 \textbf{Correlation} & PCC & $\rho$ & $\tau$ & PCC & $\rho$ & $\tau$ & PCC & $\rho$ & $\tau$ & PCC & $\rho$ & $\tau$\\
 \hline
 ROUGE-1 & 0.08 & 0.09 & 0.07 & 0.10 & 0.12 & 0.09 & 0.04 & 0.05 & 0.04 & 0.10 & 0.11 & 0.09\\
 \hline 
 ROUGE-2 & 0.05 & 0.08 & 0.07 & 0.05 & 0.10 & 0.07 & 0.03 & 0.07 & 0.05 & 0.06 & 0.09 & 0.07\\
 \hline
 ROUGE-L & 0.00* & 0.01* & 0.01* & 0.00* & 0.02* & 0.01* & -0.03 & -0.01* & -0.01* & 0.02* & 0.04 & 0.03\\
 \hline
 BLEU-1 & 0.08 & 0.08 & 0.06 & 0.14 & 0.14 & 0.11 & 0.00* & 0.03* & 0.02* & 0.08 & 0.11 & 0.09\\
 \hline
 BLEU-2 & 0.06 & 0.06 & 0.04 & 0.11 & 0.11 & 0.08 & 0.03* & 0.04* & 0.03* & 0.09 & 0.10 & 0.07\\
 \hline
 BLEU-3 & 0.08 & 0.06 & 0.05 & 0.09 & 0.09 & 0.06 & 0.04* & 0.03* & 0.02* & 0.09 & 0.08 & 0.06\\
 \hline
 BLEU-4 & 0.05 & 0.05 & 0.04 & 0.05 & 0.07 & 0.05 & 0.04* & 0.02* & 0.02* & 0.08 & 0.08 & 0.06\\
 \hline
 METEOR & 0.05 & 0.08 & 0.06 & 0.06 & 0.09 & 0.07 & 0.02* & 0.04 & 0.03 & 0.06 & 0.08 & 0.06\\
 \hline
 CIDEr & -0.02* & -0.03* & -0.02* & 0.01* & 0.02* & 0.02* & -0.08 & -0.10 & -0.07 & 0.00* & 0.00* & 0.00*\\
 \hline
 SPICE & -0.02* & -0.01* & -0.01* & 0.01* & 0.02* & 0.01* & -0.02* & -0.02* & -0.02* & 0.02* & 0.03* & 0.02*\\
 \hline
 BERTScore & 0.04 & 0.03 & 0.02 & 0.06 & 0.06 & 0.05 & 0.04 & 0.03 & 0.02 & 0.05 & 0.04 & 0.03\\
 \hline
 Coverage & -0.26 & -0.31 & -0.27 & -0.07 & -0.13 & -0.11 & -0.38 & -0.42 & -0.37 & -0.26 & -0.31 & -0.28\\
 \hline
\end{tabular}}
\vspace{-2mm}
\caption{\footnotesize Correlations on dev$_{O}$ between concept set size and evaluation metrics for our four baseline models (over the results from both seeds); values marked with * are statistically insignificant. PCC refers to Pearson correlation coefficient, $\rho$ to Spearman's rank correlation coefficient, and $\tau$ to Kendall rank correlation coefficient.}
\label{tab:correlations}
\end{table*}

\subsection{Qualitative Analysis}\label{sec:initial_qualitative_analysis}
We conduct a qualitative analysis of the baseline model outputs. We observe that several outputs are more like phrases than full coherent sentences, e.g. \textit{``body of water on a raft"}. Some generated texts are also missing important words, e.g. \textit{``A listening music and dancing in a dark room"} is clearly missing a noun before \textit{listening}. A large portion of generated texts are 
quite generic and bland, e.g. \textit{``Someone sits and listens to someone talk"}, while more detailed and specific statements are present in the human references. This can be seen as an instance of the noted \textit{``dull response" problem} faced by generation models \cite{du-black-2019-boosting,li2015diversity}, where they prefer safe, short, and frequent responses independent of the input.

Another issue is the way sentences are pieced together. Certain phrases in the outputs are either joined in the wrong order or with incorrect connectors, leading to sentences that appear to lack commonsense. For example, \textit{``body of water on a raft"} is illogical, and the phrases \textit{``body of water"} and \textit{``a raft"} are pieced together incorrectly. Example corrections include \textit{``body of water carrying a raft"} and \textit{``a raft on a body of water"}. The first changes the adverb \textit{on} joining them to the verb \textit{carrying}, and the second pieces them together in the opposite order. A similar issue occurs with the \textit{\{horse, carriage, draw\}} example in Table \ref{tab:qualitative_baselines}.


Some major takeaways are that many generations are: 1) phrases rather than full sentences and 2) poorly pieced together and lack fluency and logic compared to human references. This motivates our second set of improvements, which involves recombining extracted phrases from baseline generations into hopefully more fluent and logical sentences. This is discussed more in \S\ref{sec:phrase_recombination}.



\begin{table*}[t]
\centering
\small
\begin{tabular}{ |c|c|c| } 
 \hline
 \textbf{Concept Set} & \textbf{Baseline Generation} & \textbf{Human Reference} \\
 \hline
 \{horse, carriage, draw\} & horse drawn in a carriage & The carriage is drawn by the horse.\\
 \hline
 \{fish, catch, pole\} & fish caught on a pole & The man used a fishing pole to catch fish.\\
 \hline
 \multirow{2}{*}{\{listen, talk, sit\}} & \multirow{2}{*}{Someone sits and listens to someone talk.} & The man told the boy to sit down \\
 & & and listen to him talk.\\
 \hline
 \multirow{2}{*}{\{bathtub, bath, dog, give\}} & \multirow{2}{*}{A dog giving a bath in a bathtub.} & The teenager made a big mess in the \\
 & & bathtub giving her dog a bath.\\
 \hline
\end{tabular}
\vspace{-2mm}
\caption{\footnotesize Example generations from our baseline models versus human references.}
\label{tab:qualitative_baselines}
\end{table*}

\section{SAPPHIRE Methodology}
\label{sec:methodology}
\subsection{Concept Set Augmentation}\label{sec:concept_set_aug}
The first set of improvements is concept set augmentation, which involves augmenting the input concept sets. We try augmentation using up to 1 to 5 additional words, and train-time augmentation both with and without test-time augmentation. We observed that test-time augmentation resulted in inconsistent results that were not as effective, 
and stick with train-time only augmentation. During training, rather than feeding in the original concept sets as inputs, we instead feed in these augmented concept sets which consist of more words. The expected outputs are the same human references. During test-time, we simply feed in the original concept sets (without augmentation) as inputs.

\subsubsection{Keyword-based Augmentation}\label{sec:kw-aug}
The first type of augmentation we try is keyword-based, or \textit{Kw-aug}. We augment the train$_{CG}$ concept sets with keywords extracted from the human references using KeyBERT\footnote{\url{https://github.com/MaartenGr/KeyBERT}} \cite{grootendorst2020keybert}.  
We calculate the average semantic similarity (using cosine similarity of BERT embeddings) of the candidate keywords to the original concept set. At each stage of augmentation, we add the remaining candidate with the highest similarity.\footnote{We also tried using the least semantically similar keywords, but results were noticeably worse.} 
Some augmentation examples can be found in Table \ref{tab:augmented_concept_sets}. We train our BART and T5 models using these augmented sets, and call the resulting models BART-base-KW, BART-large-KW, T5-base-KW, and T5-large-KW.

\begin{table}[t]
\centering
\small
\resizebox{\columnwidth}{!}{\begin{tabular}{ |c|c|c| } 
 \hline
 \textbf{Method} & \textbf{Original Concept Set} & \textbf{Added Words} \\
 \hline
 Kw-aug & \{match, stadium, watch\} & \{soccer, league, fans\} \\
 \hline
 Kw-aug & \{family, time, spend\} & \{holidays\} \\
 \hline
 Kw-aug & \{head, skier, slope\} & \{cabin\} \\
 \hline
 Att-aug & \{boat, lake, drive\} & \{fisherman\} \\
 \hline
 Att-aug & \{family, time, spend\} & \{at, holidays\} \\
 \hline
 Att-aug & \{player, match, look\} & \{tennis, on, during\} \\
 \hline
\end{tabular}}
\vspace{-2mm}
\caption{ \footnotesize Example train$_{CG}$ concept set augmentations.}
\label{tab:augmented_concept_sets}
\end{table}


\subsubsection{Attention-based Augmentation}\label{sec:att-aug}
We also try attention-based augmentation, or \textit{Att-aug}. We augment the train$_{CG}$ concept sets with the words that have been most attended upon in aggregate by the other words in the human references. We pass the reference texts through BERT 
and return the attention weights at the last layer. At each stage of augmentation, we add the remaining candidate word with the highest attention. Adding the least attended words would not be effective as many are words with little meaning (e.g. simple articles such as \textit{``a"} and {``the"}). Some augmentation examples can be found in Table \ref{tab:augmented_concept_sets}. We train our BART and T5 models using these augmented sets, and call the resulting models BART-base-Att, BART-large-Att, T5-base-Att, and T5-large-Att.



\subsection{Phrase Recombination}\label{sec:phrase_recombination}
The second set of improvements is phrase recombination, which involves breaking down sentences into phrases and reconstructing them into new sentences which are hopefully more logical and coherent. For training, we use YAKE \cite{Campos2018YAKECA} to break down the train$_{CG}$ human references into phrases of up to 2, 3, and 5 n-grams long, and ensure extracted phrases have as little overlap as possible. 
During validation and testing, since we assume no access to ground-truth human references, we instead use YAKE to extract keyphrases from our baseline model generations.

We ignore extracted 1-grams as this approach focuses on phrases. 
We find words from the original concept set which are not covered by our extracted keyphrases 
and include them to ensure that coverage is maintained. 
Essentially, we form a new concept set which can also consist of phrases. Some examples can be found in Table \ref{tab:example_keyphrases}.

\begin{table*}[t]
\centering
\small
\begin{tabular}{ |c|c|c| } 
\hline
 \textbf{Original Text} & \textbf{Extracted Keyphrases} & \textbf{New Input Concept Set}\\
 \hline
 A dog wags his tail at the boy. & dog wags his tail & \{dog wags his tail\}\\
 \hline 
 hanging a painting on a wall at home & hanging a painting & \{hanging a painting, wall\}\\
 \hline
 a herd of many sheep crowded together in a stable & \multirow{2}{*}{herd of many sheep crowded} & \{herd of many sheep crowded,\\
 waiting to be dipped for ticks and other pests & &  dip, waiting\}\\
 \hline
 a soldier takes a knee while providing security & knee while providing security, & \{knee while providing security,\\
 during a patrol outside of the village. & patrol outside of the village & patrol outside of the village, take\}\\
 \hline
\end{tabular}
\vspace{-2mm}
\caption{\small Example keyphrases (up to 5-grams) extracted using YAKE from human-written training references.}
\label{tab:example_keyphrases}
\end{table*}



\subsubsection{Phrase-to-text (P2T)}\label{sec:P2T}
To piece the phrases back together, we try phrase-to-text (P2T) generation by training BART and T5 to generate full sentences given our new input sets, and call these models BART-base-P2T, BART-large-P2T, T5-base-P2T, and T5-large-P2T. During training, we choose a single random permutation of each training input set (consisting of extracted keyphrases from the human references), with the elements separated by \textit{$<$s$>$}, and the human references as the outputs. This is in order for the models to learn to be order-agnostic, which is important as one desired property of phrase recombination is the ability to combine phrases in different orders, 
as motivated by the qualitative analysis in \S\ref{sec:initial_qualitative_analysis}. During inference or test-time, we feed in a single random permutation of each test-time input set, consisting of extracted keyphrases from the corresponding baseline model's outputs.


\subsubsection{Mask Infilling (MI)}\label{sec:mask_infilling}
This method interpolates text between test-time input set elements with no training required. For example, given a test-time input set \{$c_{1}$,$c_{2}$\}, 
we feed in \textit{``$<$mask$>$ $c_{1}$ $<$mask$>$ $c_{2}$ $<$mask$>$"} and \textit{``$<$mask$>$ $c_{2}$ $<$mask$>$ $c_{1}$ $<$mask$>$"} to an MI model to fill the $<$mask$>$ tokens with text. We use BART-base and BART-large for MI, and call the approaches BART-base-MI and BART-large-MI, respectively. We use BART-base-MI on input sets consisting of extracted keyphrases from BART-base-BL and T5-base-BL, and BART-large-MI on input sets consisting of extracted keyphrases from BART-large-BL and T5-large-BL. We also try MI on the original concept sets (with no phrases).

One difficulty is determining the right input set permutation. Many contain $\geq$5 elements (meaning $\geq$5!=120 permutations), making exhaustive MI infeasible. Order of elements for infilling can result in vastly different outputs (see \S\ref{sec:post_qualitative_analysis}), as certain orders are more natural. Humans perform their own intuitive reordering of given inputs when writing, and the baselines and other approaches (e.g. Kw-aug, P2T) learn to mainly be order agnostic.

We use perplexity (PPL) from GPT-2 \cite{radford2019language} to pick the \textit{``best"} permutations for MI. We feed up to 120 permutations of each input set (with elements separated by spaces) to GPT-2 to extract their PPL, and keep the 10 with lowest PPL per example. This is not a perfect approach, but is likely better than random sampling. For each example, we perform MI on these ten permutations, and select the output with lowest GPT-2 PPL.

We found BART-large-MI outputs contain URLs, news agency names in brackets, etc. Hence, we post-process 
before output selection and evaluation. BART-base-MI does not do this. One possible explanation is that BART-large may have been pretrained on more social media and news data.



\section{Experiments}
\label{sec:experiments}

\subsection{Model Training and Selection}
For training Kw-aug, Att-aug, and P2T models, we follow baseline hyperparameters, barring learning rate (LR) which is tuned per-method. We train two seeds per model. See Appendix \ref{appendix:model_training_finetuning_details} for more.

For each model, we choose the epoch corresponding to highest ROUGE-2 on the dev split, and use beam search for decoding. The dev and test splits are different. For Kw-aug and Att-aug models, the splits are simply dev$_{CG}$ and test$_{CG}$ (or test$_{O}$), as we do not perform test-time augmentation. For P2T, the splits are dev$_{CG}$ and test$_{CG}$ (or test$_{O}$) but with the input sets replaced with new ones that include keyphrases extracted from the corresponding baseline model's outputs. 

The number of words to augment for Kw-aug and Att-aug (from 1 to 5) and maximum n-gram length of extracted keyphrases for P2T (2, 3, or 5) are hyperparameters. While we train separate versions of each model corresponding to different values of these, the final chosen model per method and model combination (such as BART-base-KW) is the one corresponding to the hyperparameter value that performs best on the dev split when averaged over both seeds. 
For MI, which involves no training, we select the variation (MI on the original concept set or new input sets with keyphrases up to 2, 3, or 5 n-grams) per model that performs best on the dev split, and only perform infilling using extracted keyphrases from the first seed baseline generations. These are the selected models we report the test$_{CG}$ and test$_{O}$ results of in \S\ref{sec:results_and_analysis}.



\subsection{Human Evaluation}
We ask annotators to evaluate 48 test$_{CG}$ examples from the human references, baseline outputs, 
and various method (excluding MI) outputs for BART-large and T5-base. We choose these two as they cover both model types and sizes, and exclude MI as it performs noticeably worse on the automatic evaluation (see \S\ref{sec:automatic_eval}). See Appendix \S\ref{sec:appendix_human_eval} for more.

The annotators evaluate fluency and commonsense of the texts on 1-5 scales. Fluency, also known as naturalness, is a measure of how human-like a text is. Commonsense is the plausibility of the events described. We do not evaluate coverage as automatic metrics suffice; coverage is more objective compared to fluency and commonsense.

\begin{table*}[!htbp]
\centering
\small
\begin{tabular}{ |c|c|c|c|c|c| }
\hline
 & \multicolumn{5}{c|}{\textbf{BART-base}}\\
 \hline
 \textbf{Metrics\textbackslash Methods}& Baseline & Kw-aug & Att-aug & P2T & BART-base-MI\\
 \hline
 ROUGE-1 & 43.96$\pm0.03$ & \textbf{45.01}$\pm0.00$ & 44.99$\pm0.10$ & 44.87$\pm0.42$ & 44.83\\
 \hline 
 ROUGE-2 & 17.31$\pm0.02$ & \textbf{18.33}$\pm0.06$ & 18.18$\pm0.04$ & 18.04$\pm0.13$ & 17.44\\
 \hline
 ROUGE-L & 36.65$\pm0.00$ & 37.28$\pm0.24$ & \textbf{37.76}$\pm0.12$ & 37.28$\pm0.11$ & 34.47\\
 \hline
 BLEU-1 & \textbf{73.20}$\pm0.28$ & 73.00$\pm0.85$ & 73.00$\pm0.14$ & 73.15$\pm1.06$ & 69.90\\
 \hline
 BLEU-2 & 54.50$\pm0.14$ & 55.35$\pm0.49$ & \textbf{55.70}$\pm0.28$ & 55.65$\pm0.35$ & 49.00\\
 \hline
 BLEU-3 & 40.40$\pm0.14$ & 41.35$\pm0.21$ & 41.40$\pm0.28$ & \textbf{41.85}$\pm0.35$ & 34.70\\
 \hline
 BLEU-4 & 30.10$\pm0.14$ & 31.10$\pm0.14$ & 30.95$\pm0.07$ & \textbf{31.75}$\pm0.35$ & 24.70\\
 \hline
 METEOR & 30.35$\pm0.35$ & 30.50$\pm0.28$ & 30.70$\pm0.14$ & \textbf{31.05}$\pm0.49$ & 29.70\\
 \hline
 CIDEr & 15.56$\pm0.10$ & \textbf{16.18}$\pm0.12$ & 15.68$\pm0.00$ & 16.14$\pm0.33$ & 14.43\\
 \hline
 SPICE & 30.05$\pm0.07$ & 30.45$\pm0.07$ & 30.65$\pm0.35$ & \textbf{30.95}$\pm0.21$ & 28.40\\
 \hline
 BERTScore & 59.19$\pm0.32$ & 59.32$\pm0.25$ & \textbf{59.72}$\pm0.03$ & 59.54$\pm0.05$ & 53.73\\
 \hline
 Coverage & 90.43$\pm0.17$ & 91.44$\pm0.95$ & 91.23$\pm0.21$ & 91.47$\pm2.93$ & \textbf{96.23}\\
 \hline
\end{tabular}
\vspace{-2mm}
\caption{\footnotesize Automatic evaluation results (with standard deviations) for BART-base on test$_{CG}$, averaged over two seeds for trained models. Bold corresponds to best performance on that metric.}
\vspace{-1mm}
\label{tab:automatic_results_BART-base}
\end{table*}

\begin{table*}[!htbp]
\centering
\small
\begin{tabular}{ |c|c|c|c|c|c| }
\hline
 & \multicolumn{5}{c|}{\textbf{BART-large}}\\
 \hline
 \textbf{Metrics\textbackslash Methods} & Baseline & Kw-aug & Att-aug & P2T & BART-large-MI\\
 \hline
 ROUGE-1 & 45.67$\pm0.25$ & 46.71$\pm0.05$ & \textbf{46.78}$\pm0.14$ & 46.26$\pm0.29$ & 41.69\\
 \hline 
 ROUGE-2 & 18.77$\pm0.04$ & 19.64$\pm0.05$ & \textbf{19.92}$\pm0.19$ & 19.37$\pm0.17$ & 15.40\\
 \hline
 ROUGE-L & 37.83$\pm0.29$ & 38.38$\pm0.01$ & \textbf{38.53}$\pm0.03$ & 38.22$\pm0.16$ & 33.32\\
 \hline
 BLEU-1 & 74.45$\pm0.21$ & 76.20$\pm0.99$ & 76.55$\pm0.92$ & \textbf{77.10}$\pm0.85$ & 63.90\\
 \hline
 BLEU-2 & 56.25$\pm0.78$ & 58.60$\pm0.14$ & \textbf{59.60}$\pm0.00$ & 58.95$\pm0.64$ & 42.40\\
 \hline
 BLEU-3 & 42.15$\pm0.49$ & 44.00$\pm0.00$ & \textbf{45.20}$\pm0.42$ & 44.70$\pm0.14$ & 29.20\\
 \hline
 BLEU-4 & 32.10$\pm0.42$ & 33.40$\pm0.28$ & \textbf{34.50}$\pm0.42$ & 34.25$\pm0.21$ & 20.50\\
 \hline
 METEOR & 31.70$\pm0.14$ & 32.60$\pm0.57$ & 32.65$\pm0.49$ & \textbf{33.00}$\pm0.14$ & 28.30\\
 \hline
 CIDEr & 16.42$\pm0.09$ & 17.37$\pm0.08$ & 17.49$\pm0.49$ & \textbf{17.50}$\pm0.02$ & 12.32\\
 \hline
 SPICE & 31.85$\pm0.21$ & 33.15$\pm0.49$ & 33.30$\pm0.28$ & \textbf{33.60}$\pm0.00$ & 26.10\\
 \hline
 BERTScore & 59.95$\pm0.29$ & 60.83$\pm0.29$ & 60.87$\pm0.45$ & \textbf{61.30}$\pm0.66$ & 48.56\\
 \hline
 Coverage & 94.49$\pm0.53$ & 96.74$\pm1.20$ & 96.02$\pm1.17$ & \textbf{97.02}$\pm0.15$ & 95.33\\
 \hline
\end{tabular}
\vspace{-2mm}
\caption{\footnotesize Automatic evaluation results (with standard deviations) for BART-large on test$_{CG}$, averaged over two seeds for trained models. Bold corresponds to best performance on that metric.}
\vspace{-1mm}
\label{tab:automatic_results_BART-large}
\end{table*}

\begin{table*}[!htbp]
\centering
\small
\begin{tabular}{ |c|c|c|c|c|c| }
\hline
 & \multicolumn{5}{c|}{\textbf{T5-base}}\\
 \hline
 \textbf{Metrics\textbackslash Methods} & Baseline & Kw-aug & Att-aug & P2T & BART-base-MI\\
 \hline
 ROUGE-1 & 44.63$\pm0.13$ & 46.42$\pm0.01$ & \textbf{46.75}$\pm$0.11 & 45.73$\pm0.27$ & 44.92\\
 \hline 
 ROUGE-2 & 18.40$\pm0.14$ & \textbf{19.36}$\pm0.24$ & 19.20$\pm$0.17 & 18.51$\pm0.11$ & 17.98\\
 \hline
 ROUGE-L & 37.60$\pm0.16$ & \textbf{38.68}$\pm0.08$ & 38.51$\pm$0.21 & 38.07$\pm0.10$ & 34.88\\
 \hline
 BLEU-1 & 73.60$\pm0.85$ & \textbf{76.25}$\pm0.35$ & 76.00$\pm$0.28 & 75.65$\pm1.20$ & 70.20\\
 \hline
 BLEU-2 & 57.00$\pm0.71$ & \textbf{59.55}$\pm0.64$ & 58.75$\pm$0.35 & 58.15$\pm0.64$ & 50.50\\
 \hline
 BLEU-3 & 42.75$\pm0.49$ & \textbf{45.10}$\pm0.85$ & 44.00$\pm$0.28 & 43.45$\pm0.07$ & 36.20\\
 \hline
 BLEU-4 & 32.70$\pm0.42$ & \textbf{34.45}$\pm0.92$ & 33.30$\pm$0.28 & 33.10$\pm0.28$ & 26.10\\
 \hline
 METEOR & 31.05$\pm0.49$ & 31.85$\pm0.07$ & 31.90$\pm$0.14 & \textbf{32.05}$\pm0.35$ & 30.20\\
 \hline
 CIDEr & 16.26$\pm0.25$ & \textbf{17.37}$\pm0.04$ & 17.04$\pm$0.21 & 16.84$\pm0.11$ & 14.83\\
 \hline
 SPICE & 31.95$\pm0.07$ & 32.75$\pm0.21$ & 32.85$\pm$0.21 & \textbf{33.20$\pm0.14$} & 29.70\\
 \hline
 BERTScore & 61.40$\pm0.34$ & \textbf{61.88}$\pm0.06$ & 61.28$\pm$0.10 & 61.46$\pm0.01$ & 55.04\\
 \hline
 Coverage & 90.96$\pm1.77$ & 94.92$\pm0.45$ & 96.00$\pm$0.03 & 94.78$\pm0.83$ & \textbf{96.03}\\
 \hline
\end{tabular}
\vspace{-2mm}
\caption{\footnotesize Automatic evaluation results (with standard deviations) for T5-base on test$_{CG}$, averaged over two seeds for trained models. Bold corresponds to best performance on that metric.}
\label{tab:automatic_results_T5-base}
\vspace{-1mm}
\end{table*}

\begin{table*}[!htbp]
\centering
\small
\begin{tabular}{ |c|c|c|c|c|c| }
\hline
 & \multicolumn{5}{c|}{\textbf{T5-large}}\\
 \hline
 \textbf{Metrics\textbackslash Methods} & Baseline & Kw-aug & Att-aug & P2T & BART-large-MI\\
 \hline
 ROUGE-1 & 46.26$\pm0.17$ & \textbf{47.47}$\pm0.16$ & 47.40$\pm0.12$ & 46.72$\pm0.26$ & 42.78\\
 \hline 
 ROUGE-2 & 19.62$\pm0.17$ & 20.02$\pm0.07$ & \textbf{20.19}$\pm0.01$ & 19.76$\pm0.22$ & 16.61\\
 \hline
 ROUGE-L & 39.21$\pm0.22$ & 39.84$\pm0.12$ & \textbf{39.97}$\pm0.06$ & 39.19$\pm0.09$ & 34.52\\
 \hline
 BLEU-1 & 77.45$\pm0.21$ & 78.70$\pm0.28$ & \textbf{78.95}$\pm0.07$ & 77.90$\pm0.57$ & 66.80\\
 \hline
 BLEU-2 & 60.75$\pm0.21$ & 62.10$\pm0.14$ & \textbf{62.35}$\pm0.07$ & 61.00$\pm0.42$ & 45.90\\
 \hline
 BLEU-3 & 46.60$\pm0.14$ & 47.65$\pm0.21$ & \textbf{47.95}$\pm0.21$ & 46.75$\pm0.49$ & 32.70\\
 \hline
 BLEU-4 & 36.30$\pm0.00$ & 36.80$\pm0.28$ & \textbf{37.35}$\pm0.49$ & 36.10$\pm0.42$ & 23.90\\
 \hline
 METEOR & 33.30$\pm0.14$ & 33.55$\pm0.07$ & \textbf{33.70}$\pm0.00$ & 33.35$\pm0.21$ & 29.10\\
 \hline
 CIDEr & 17.90$\pm0.15$ & 18.40$\pm0.18$ & \textbf{18.43}$\pm0.10$ & 17.89$\pm0.08$ & 13.34\\
 \hline
 SPICE & 34.25$\pm0.07$ & \textbf{34.50}$\pm0.28$ & 33.70$\pm0.14$ & 34.00$\pm0.28$ & 28.00\\
 \hline
 BERTScore & 62.65$\pm0.07$ & \textbf{62.91}$\pm0.15$ & 62.78$\pm0.21$ & 62.46$\pm0.11$ & 50.57\\
 \hline
 Coverage & 94.23$\pm0.21$ & 95.92$\pm0.02$ & \textbf{96.08}$\pm0.09$ & 95.44$\pm0.58$ & 96.03\\
 \hline
\end{tabular}
\vspace{-2mm}
\caption{ \footnotesize Automatic evaluation results (with standard deviations) for T5-large on test$_{CG}$, averaged over two seeds for trained models. Bold corresponds to best performance on that metric.}
\label{tab:automatic_results_T5-large}
\vspace{-2mm}
\end{table*}

\section{Results and Analysis}
\label{sec:results_and_analysis}

\begin{table}[!htbp]
\centering
\resizebox{\columnwidth}{!}{\begin{tabular}{ |l|c|c|c|c|c|c|}
\hline
 & \multicolumn{1}{c|}{\textbf{BART-base}} & \multicolumn{2}{c|}{\textbf{BART-large}} & \multicolumn{1}{c|}{\textbf{T5-base}} & \multicolumn{1}{c|}{\textbf{T5-large}}\\
 \hline
 \textbf{p-values} & P2T & Att-aug & P2T & Kw-aug & Att-aug\\
 \hline
 ROUGE-1 & 1.58E-05 & 1.58E-05 & 7.58E-04 & 1.58E-05 & 1.58E-05\\
 \hline 
 ROUGE-2 & 6.32E-05 & 1.58E-05 & 2.18E-03 & 1.58E-05 & 2.20E-03\\
 \hline
 ROUGE-L & 6.32E-05 & 8.53E-04 & 2.78E-02 & 1.58E-05 & 1.58E-05\\
 \hline
 BLEU-1 & \textbf{3.63E-01} & 1.39E-04 & 6.94E-05 & 6.94E-05 & 1.11E-03\\
 \hline
 BLEU-2 & 1.11E-03 & 6.94E-05 & 6.94E-05 & 6.94E-05 & 5.69E-03\\
 \hline
 BLEU-3 & 3.26E-02 & 1.04E-03 & 9.03E-04 & 4.17E-04 & 3.40E-02\\
 \hline
 BLEU-4 & \textbf{5.68E-02} & \textbf{1.57E-01} & 8.40E-03 & 1.83E-02 & \textbf{2.66E-01}\\
 \hline
 METEOR & 1.57E-02 & 9.03E-04 & 6.94E-05 & 2.08E-04 & \textbf{7.27E-01}\\
 \hline
 CIDEr & 6.25E-04 & 2.08E-04 & 6.94E-05 & 6.94E-05 & 5.07E-03\\
 \hline
 SPICE & 1.53E-03 & 6.25E-04 & 6.94E-05 & 1.43E-02 & \textbf{9.16E-01}\\
 \hline
 BERTScore & 3.33E-03 & 1.58E-05 & 1.58E-05 & 1.58E-05 & \textbf{1.42E-01}\\
 \hline
 Coverage & 3.16E-05 & 1.58E-05 & 1.58E-05 & 1.58E-05 & 1.58E-05\\
 \hline
\end{tabular}}
\vspace{-2mm}
\caption{\small Statistical significance p-values (from Pitman's permutation tests) for the best performing method(s) per model compared to the corresponding baselines. Insignificant p-values (using $\alpha=0.05$ or 5E-02) are bolded.}
\label{tab:stat_sig}
\end{table}

\begin{table*}[t]
\centering
\small
\begin{tabular}{ |c|cc|cc|c|cc|c| }
\hline
 \textbf{Models\textbackslash Metrics} & \multicolumn{2}{c|}{ROUGE-2/L} & \multicolumn{2}{c|}{\textbf{BLEU}-3/\textbf{4}} & METEOR & \textbf{CIDEr} & \textbf{SPICE} & Coverage\\
 \hline
 T5-base (reported baseline) & 14.63 & 34.56 & 28.76 & 18.54 & 23.94 & 9.40 & 19.87 & 76.67 \\
 \hline
 BART-large (reported baseline) & 22.02 & 41.78 & 39.52 & 29.01 & 31.83 & 13.98 & 28.00 & 97.35 \\
 \hline
 T5-large (reported baseline) & 21.74 & 42.75 & 43.01 & 31.96 & 31.12 & 15.13 & 28.86 & 95.29 \\
 \Xhline{2\arrayrulewidth}
 EKI-BART \cite{fan2020enhanced} & - & - & - & 35.945 & - & \underline{16.999} & 29.583 & - \\
 \hline
 KG-BART \cite{liu2020kg} & - & - & - & 33.867 & - & 16.927 & 29.634 & - \\
 \hline
 RE-T5 \cite{wang2021retrieval} & - & - & - & \textbf{40.863} & - & \textbf{17.663} & \textbf{31.079} & - \\
 \Xhline{2\arrayrulewidth}
 BART-base-P2T & 20.83 & 42.91 & 40.74 & 29.918 & 30.61 & 14.670 & 26.960 & 92.84\\ 
 \hline
 T5-base-P2T & 22.38 & 44.59 & 44.97 & 33.577 & 31.95 & 16.152 & 29.104 & 95.55\\ 
 \hline
 BART-large-KW & 22.25 & 43.38 & 43.87 & 32.956 & 32.26 & 16.065 & 28.335 & 96.16\\ 
 \hline
 BART-large-Att & 22.22 & 43.80 & 44.61 & 33.405 & 32.03 & 16.036 & 28.452 & 96.43\\ 
 \hline
 BART-large-P2T & 22.65 & 43.84 & 44.78 & 33.961 & 32.18 & 16.174 & 28.462 & 96.20\\ 
 \hline
 T5-large-KW & 23.79 & 45.73 & 48.06 & 37.023 & 32.85 & 16.987 & 29.659 & 95.32\\ 
 \hline 
 T5-large-Att & 23.94 & 45.87 & 47.99 & 36.947 & 32.79 & 16.943 & 29.607 & 95.43\\ 
 \hline 
 T5-large-P2T & 23.89 & 45.77 & 48.08 & \underline{37.119} & 32.94 & 16.901 & \underline{29.751} & 94.82\\ 
 \hline 
\end{tabular}
\vspace{-2mm}
\caption{\footnotesize Automatic evaluation results of select SAPPHIRE models on test$_{O}$ (evaluated by the CommonGen authors). For BART-base and T5-base, we report the best SAPPHIRE model on test$_{O}$ (P2T), and all three models for BART-large and T5-large. We compare to \citet{lin-etal-2020-commongen}'s reported baseline numbers, noting that they did not report BART-base, and published models on their leaderboard\footnoteref{leaderboard} that outperform the baselines at the time of writing. Bold corresponds to best performance (for BLEU-4, CIDEr, and SPICE, since their leaderboard only reports these three), and underline corresponds to second best performance.}
\label{tab:results_original_test_split}
\end{table*}

\begin{table}[!htbp]
\centering
\small
\resizebox{\columnwidth}{!}{
\begin{tabular}{ |c|c|c|c|}
\hline
\underline{Model} & \underline{Method} & \textbf{Fluency} & \textbf{Commonsense} \\
 \hline
 \multirow{4}{*}{\textbf{BART-large}} & Baseline & 3.92 & 4.06\\
 \cline{2-4}
 & Kw-aug & 4.13 & 3.92\\
 \cline{2-4}
 & Att-aug & 4.10 & 4.06\\
 \cline{2-4}
 & P2T & \textbf{4.17} & \textbf{4.13}\\
 \hline
 \multirow{4}{*}{\textbf{T5-base}} & Baseline & 4.02 & 3.83\\
 \cline{2-4}
 & Kw-aug & 4.04 & 4.04\\
 \cline{2-4}
 & Att-aug & \textbf{4.13} & 3.98\\
 \cline{2-4}
 & P2T & 4.02 & \textbf{4.08}\\
 \hline
\multicolumn{2}{|c|}{\textbf{Human}} & 4.14 & 4.32\\
\hline
\end{tabular}}
\vspace{-2mm}
\caption{\small Avg. human eval results on test$_{CG}$, rated on 1-5 scales. Bold corresponds to best performance for that model.}
\label{tab:human_eval_results}
\end{table}

Automatic evaluation results on test$_{CG}$ can be found in Tables \ref{tab:automatic_results_BART-base}, \ref{tab:automatic_results_BART-large}, \ref{tab:automatic_results_T5-base}, \ref{tab:automatic_results_T5-large}, and results on test$_{O}$ in Table \ref{tab:results_original_test_split}. Human evaluation results on test$_{CG}$ can be found in Table~\ref{tab:human_eval_results}. Single keyword augmentation performs best for Kw-aug across models. Two word augmentation performs best for Att-aug, except T5-base where three word augmentation performs best. Keyphrases up to 2-grams long perform best for P2T, except T5-large where 3-grams perform best. All models perform best with keyphrases up to 5-grams long for MI. These are the results reported here, and graphs displaying other hyperparameter results on test$_{CG}$ are in Appendix \ref{appendix:hyperparam_graphs}. Table \ref{tab:qualitative_body} contains qualitative examples, and more can be found in Appendix \S\ref{sec:appendix_qualitative_examples}.

\subsection{Automatic Evaluation}\label{sec:automatic_eval}
We see from Tables \ref{tab:automatic_results_BART-base} to \ref{tab:automatic_results_T5-large} that SAPPHIRE methods outperform the baselines on most/all metrics across the models on test$_{CG}$. The only exception is MI, which performs worse other than coverage.

For BART-base, Kw-aug, Att-aug, and P2T all outperform the baseline across the metrics. 
For BART-large, Att-aug and P2T outperform the baseline heaviest, with noticeable increases to all metrics. For T5-base, all methods outperform the baseline, with Kw-aug performing best. Att-aug performs best for T5-large, and SAPPHIRE appears relatively less effective for T5-large. T5-large is the strongest baseline, and hence further improving its performance is possibly more difficult.

MI performs worse across most metrics except coverage, likely as MI almost always keeps inputs intact in their exact form. This is however possibly one reason for its low performance; it is less flexible. Further, as discussed in \S\ref{sec:mask_infilling}, MI is highly dependent on the input order. See \S\ref{sec:post_qualitative_analysis} for more.

Table \ref{tab:stat_sig} contains statistical significance p-values from Pitman's permutation tests \cite{pitman1937significance} 
for what we adjudged to be the best performing method(s) per model compared to corresponding baselines on test$_{CG}$. Most metrics across the methods are significant 
compared to the baselines. 

From Table \ref{tab:results_original_test_split}, we see that SAPPHIRE models outperform the corresponding baselines reported in \citet{lin-etal-2020-commongen} on test$_O$. 
T5-large-KW and P2T outperform EKI-BART \cite{fan2020enhanced} and KG-BART \cite{liu2020kg} on both SPICE and BLEU-4, which are two SOTA published CommonGen models that use external knowledge from corpora and KGs. As SPICE is used to rank the CommonGen leaderboard\footnote{\label{leaderboard}\url{https://inklab.usc.edu/CommonGen/leaderboard.html}}, T5-large-KW and P2T would place highly. SAPPHIRE models do lag behind the SOTA published RE-T5 \cite{wang2021retrieval}, showing potential for further improvement. Further, the BART-large SAPPHIRE models perform worse than EKI-BART and KG-BART, but not by a substantial margin. We emphasize again that SAPPHIRE simply uses the data itself and the baseline generations, rather than external knowledge. Hence, SAPPHIRE's performance gains over the baselines and certain SAPPHIRE models matching or outperforming SOTA models that leverage external information is quite impressive.


\subsection{Human Evaluation}\label{sec:human_eval_analysis}
Table~\ref{tab:human_eval_results} shows human evaluation results on test$_{CG}$ for human references and 
methods (excluding MI) using BART-large and T5-base. SAPPHIRE generally outperforms the baselines. BART-large-P2T performs noticeably higher on both fluency and commonsense. 
For T5-base, all three methods outperform the baseline across both metrics. 
Compared to humans, our best methods have comparable fluency, but still lag noticeably on commonsense, demonstrating that 
human-level generative commonsense reasoning is indeed challenging.

\subsection{Qualitative Analysis}\label{sec:post_qualitative_analysis}
We see from Table \ref{tab:qualitative_body} that many baseline outputs contain issues found in \S\ref{sec:initial_qualitative_analysis}, e.g. incomplete or illogical sentences. Human references are fluent, logical, and sometimes more creative (e.g. example 5), which all methods still lack in comparison.

For example 1, the baseline generation \textit{``hands sitting on a chair"} misses the concept \textit{``toy"}, whereas our methods do not. Kw-aug and Att-aug output complete and logical sentences. For example 2, the baseline generation of \textit{``a camel rides a camel"} is illogical. Our methods output more logical and specific sentences. For example 3, our methods generate more complete and coherent sentences than the baseline, which lacks a subject (does not mention \textit{who} is \textit{``walking"}). For example 4, the baseline generation \textit{``bus sits on the tracks"} is illogical as buses park on roads. Our methods do not suffer from this and output more reasonable text. For example 5, the baseline generation \textit{``A lady sits in a sunglass."} is completely illogical. Kw-aug, Att-aug, and P2T all output logical text. For example 6, the baseline output \textit{``Someone stands in front of someone holding a hand"} is generic and bland. 
Kw-aug, Att-aug, and P2T all output more specific and detailed text rather than simply referring to \textit{``someone"}. Overall, SAPPHIRE generates text that is more complete, fluent, logical, and with greater coverage, addressing many baseline issues (\S\ref{sec:initial_qualitative_analysis}).

However, SAPPHIRE methods are imperfect. P2T relies heavily on the original generation. For example 1, the baseline output \textit{``hands sitting on a chair"} is extracted as a keyphrase, and used in the P2T output \textit{``hands sitting on a chair with toys"}. While coverage improves, the text is still illogical. For example 2, P2T still misses the \textit{``walk"} concept. While the Att-aug output of \textit{``A man is riding camel as he walks through the desert."} is more logical than the baseline's, it is still not entirely logical as the man cannot ride the camel and walk at the same time. MI outputs logical and fluent text for examples 2 and 3. 
For the other examples, 
the generated texts are illogical, not fluent, or incomplete. 

This is likely due to input permutation having a strong effect on output quality. 
For example, \textit{``wave"} before \textit{``falls off a surf board"} leads to an illogical output \textit{``A wave falls off a surf board."}, where the reverse order results in a more logical output \textit{``A man falls off a surf board and hits a wave."} As discussed in \S\ref{sec:mask_infilling}, our method of selecting best permutations is likely imperfect. Further, BART-MI usually does not inflect inputs and retains them in exact form, 
unlike the baselines and other methods which learn to inflect words 
(e.g. singular to plural). 
We believe BART-MI has potential if these weaknesses can be addressed.

\begin{table}[h!]
\centering
\small
\addtolength{\tabcolsep}{-4pt}
\resizebox{\columnwidth}{!}{
\begin{tabular}{|p{2.2cm}|p{7.7cm}|}
\hline 
\textbf{Method} & \textbf{Text} \\ \hline
Concept Set & \{sit, chair, toy, hand\} (example 1)\\\hline
BART-base-BL & \textcolor{red}{hands sitting on a chair} \\ \hline
BART-base-KW & {\textcolor{brown}{A boy sits on a chair with a toy in his hand.}} \\ \hline 
BART-base-Att & {\textcolor{blue}{A child sits on a chair with a toy in his hand.}} \\ \hline 
BART-base-P2T & {\textcolor{violet}{hands sitting on a chair with toys}} \\ \hline 
BART-base-MI & {\textcolor{orange}{Children's hands sit on a chair with a toy.}} \\  \hline
Human & {\textcolor{teal}{A baby sits on a chair with a toy in one of its hands.}}\\
\Xhline{3\arrayrulewidth}
Concept Set & \{camel, desert, ride, walk\} (example 2)\\\hline
BART-base-BL & \textcolor{red}{a camel rides a camel in the desert} \\ \hline
BART-base-KW & {\textcolor{brown}{A camel rides down a walkway in the desert.}} \\ \hline 
BART-base-Att & {\textcolor{blue}{A man is riding camel as he walks through the desert.}} \\ \hline 
BART-base-P2T & {\textcolor{violet}{A camel rides down a trail in the desert.}} \\ \hline 
BART-base-MI & {\textcolor{orange}{In the desert, a man rides a camel for a walk.}} \\ \hline
Human & {\textcolor{teal}{A loud group of people walk around the desert and ride camels.}} \\
\Xhline{3\arrayrulewidth}
Concept Set & \{jacket, wear, snow, walk\} (example 3)\\\hline
BART-large-BL & \textcolor{red}{walking in the snow wearing a furry jacket} \\ \hline
BART-large-KW & {\textcolor{brown}{A man wearing a jacket is walking in the snow.}} \\ \hline 
BART-large-Att & {\textcolor{blue}{A man in a blue jacket is walking in the snow.}} \\ \hline 
BART-large-P2T & {\textcolor{violet}{A man is wearing a furry jacket as he walks in the snow.}} \\ \hline 
BART-large-MI & {\textcolor{orange}{A walk in the snow wearing a furry jacket}} \\ \hline
Human & {\textcolor{teal}{A man wears a jacket and walks in the snow.}}\\
\Xhline{3\arrayrulewidth}
Concept Set & \{bench, bus, wait, sit\} (example 4)\\\hline
BART-large-BL & \textcolor{red}{A bus sits on the tracks with people waiting on benches.} \\ \hline
BART-large-KW & {\textcolor{brown}{A bus sits next to a bench waiting for passengers.}} \\ \hline 
BART-large-Att & {\textcolor{blue}{A woman sits on a bench waiting for a bus.}} \\ \hline 
BART-large-P2T & {\textcolor{violet}{A bus sits at a stop waiting for passengers to get off the bench.}} \\ \hline 
\multirow{2}{*}{BART-large-MI} & {\textcolor{orange}{There are people waiting on benches outside bus stops}} \\
& {\textcolor{orange}{to sit down. pic.twitter.}} \\ \hline
Human & {\textcolor{teal}{The man sat on the bench waiting for the bus.}}\\
\Xhline{3\arrayrulewidth}
Concept Set & \{sunglass, wear, lady, sit\} (example 5)\\\hline
T5-base-BL & \textcolor{red}{A lady sits in a sunglass.} \\ \hline
T5-base-KW & {\textcolor{brown}{A lady sits next to a man wearing sunglasses.}} \\ \hline 
T5-base-Att & {\textcolor{blue}{A lady sits wearing sunglasses.}} \\ \hline 
T5-base-P2T & {\textcolor{violet}{A lady sits next to a man wearing sunglasses.}} \\ \hline 
BART-base-MI & {\textcolor{orange}{A young lady sits in a sunglass to wear.}} \\ \hline
\multirow{2}{*}{Human} & {\textcolor{teal}{The lady wants to wear sunglasses, sit, relax,}}\\
& \textcolor{teal}{and enjoy her afternoon.} \\
\Xhline{3\arrayrulewidth}
Concept Set & \{hold, hand, stand, front\} (example 6)\\\hline
T5-large-BL & \textcolor{red}{Someone stands in front of someone holding a hand.} \\ \hline
T5-large-KW & {\textcolor{brown}{Two men stand in front of each other holding hands.}} \\ \hline
T5-large-Att & {\textcolor{blue}{A man stands in front of a woman holding a hand.}} \\ \hline 
T5-large-P2T & {\textcolor{violet}{A man standing in front of a man holding a hand.}} \\ \hline 
BART-large-MI & {\textcolor{orange}{Mr. Trump holding a hand to stand in front of}} \\ \hline
Human & {\textcolor{teal}{A man stands and holds his hands out in front of him.}} \\
\hline
\end{tabular}}
\vspace{-2mm}
\caption{\small Qualitative examples for test$_{CG}$. Color coded: \textcolor{red}{baseline}, \textcolor{brown}{Kw-aug}, \textcolor{blue}{Att-aug}, \textcolor{violet}{P2T}, \textcolor{orange}{MI}, and \textcolor{teal}{human reference}.}
\label{tab:qualitative_body}
\end{table}

\section{Related Work}
\label{sec:related_work}

\paragraph{Constrained Text Generation:}
There has been more work on constrained text generation. 
\citet{miao2019cgmh} use Metropolis-Hastings sampling to determine 
token-level edits at each step of generation. 
\citet{feng2019keep} introduce Semantic Text Exchange to adjust the semantics of a text given a \textit{replacement entity}. 
\citet{gangal2021nareor} propose narrative reordering (NAREOR) to rewrite stories in different narrative orders while preserving plot.



\paragraph{Data-to-text NLG:} A wide range of data-to-text NLG benchmarks have been proposed, e.g. for generating weather reports \cite{liang2009learning}, game commentary \cite{jhamtani2018learning}, and recipes \cite{kiddon2016globally}. E2E-NLG \cite{duvsek2018findings} and WebNLG \cite{gardent2017webnlg} are two benchmarks that involve generating text from meaning representation (MR) and triple sequences. 
\citet{montella2020denoising} use target Wiki sentences with parsed OpenIE triples as weak supervision for WebNLG. \citet{tandon2018tnt} permute input MRs to augment examples for E2E-NLG. 

\paragraph{Commonsense Reasoning and Incorporation:}
\citet{talmor2019olmpics} show that not all pretrained LMs can reason through commonsense tasks. 
Other works investigate commonsense injection into models; one popular way is through knowledge graphs (KGs). One large commonsense KG is COMET, which trains on KG edges to learn connections between words and phrases. COSMIC \cite{ghosal2020cosmic} uses COMET to inject commonsense. EKI-BART \cite{fan2020enhanced} and KG-BART \cite{liu2020kg} show that external knowledge (from corpora and KGs) can improve performance on CommonGen. Distinctly, SAPPHIRE obviates reliance on external knowledge. 

\section{Conclusion and Future Work}
\label{sec:conclusion}
In conclusion, we motivated and proposed several improvements for concept-to-text generation which we call SAPPHIRE: Set Augmentation and Post-hoc PHrase Infilling and REcombination. 
We demonstrated their effectiveness on CommonGen through experiments on BART and T5. Extensive evaluation showed that SAPPHIRE improves model performance, addresses many issues of the baselines, 
and has potential for further exploration.

Potential future work includes improving mask infilling performance, and trying combinations of SAPPHIRE techniques as they could be complementary. 
Better exploiting regularities of CommonGen-like tasks, e.g. invariance to input order, presents another avenue. SAPPHIRE methods can also be investigated for other data-to-text NLG tasks, e.g. WebNLG, and explored for applications such as 
improving the commonsense reasoning of personalized dialog agents \cite{Li_Jiang_Feng_Sprague_Zhou_Hoey_2020}, data augmentation for NLG \cite{feng-etal-2020-genaug,feng2021survey}, 
and constructing pseudo-references for long-context NLG \cite{gangal2021improving}.

\section*{Acknowledgments}
\vspace{-1mm}
We thank our anonymous reviewers, Graham Neubig, Ritam Dutt, Divyansh Kaushik, and Zhengbao Jiang for their comments and suggestions. 

\clearpage
\bibliographystyle{acl_natbib}
\bibliography{anthology,acl2021}

\clearpage
\appendix
\section*{Appendices}

\section{Model Training and Generation Details}\label{appendix:model_training_finetuning_details}

T5-large consists of 770M params, T5-base 220M params, BART-large 406M params, and BART-base 139M params. We train two seeded versions of each baseline model and SAPPHIRE model. For all models, we use beam search with a beam size of 5, decoder early stopping, a decoder length penalty of 0.6, encoder and decoder maximum lengths of 32, and a decoder minimum length of 1. For model training, we use a batch size of 128 for T5-base and BART-base, 32 for BART-large, and 16 for T5-large. For T5-base, T5-large, and BART-base, we use 400 warmup steps, and 500 for BART-large. We train all models up to a reasonable number of epochs (e.g. 10 or 20) and perform early stopping using our best judgment (e.g. if metrics have continually decreased for multiple epochs). The learning rates for SAPPHIRE models were determined by trying a range of values (e.g. from 1e-6 to 1e-4), and finding ones which led to good convergence behavior (e.g. validation metrics increase at a decently steady rate and reach max. after a reasonable number of epochs). For the best-performing models, learning rates are as follows (each set consists of \{baseline,Kw-aug,Att-aug,P2T\}): BART-base = \{3e-05,2e-05,3e-05,1e-05\}, BART-large = \{3e-05,2e-05,2e-05,5e-06\}, T5-base = \{5e-05,5e-05,5e-05,1e-05\}, T5-large = \{2e-05,2e-05,2e-05,5e-06\}.

Training was done using single RTX 2080 Ti and Titan Xp GPUs, and Google Colab instances which alternately used a single V100, P100, or Tesla T4 GPU. The vast majority of the training was done on a single V100 per model. T5-base models trained in approx. 1 hour, BART-base models in approx. 45 minutes, T5-large models in approx. 4 hours, and BART-large models in approx. 1.5-2 hours.



\section{Full Re-implementation versus Reported Model Numbers}\label{appendix:reimplementation_numbers}
See Table \ref{tab:full_reimplementation_stats} for full comparison of our re-implemented CommonGen models compared to the original reported numbers in \citet{lin-etal-2020-commongen}.

\begin{table}[ht!]
\centering
\small
\addtolength{\tabcolsep}{-4pt}
\resizebox{\columnwidth}{!}{
\begin{tabular}{|p{2.2cm}|p{7.8cm}|}
\hline 
\textbf{Method} & \textbf{Text} \\ \hline
Concept Set & \{food, eat, hand, bird\} \\\hline
BART-base-BL & \textcolor{red}{hands of a bird eating food} \\ \hline
BART-base-KW & {\textcolor{brown}{a bird eats food from a hand}} \\ \hline 
BART-base-Att & {\textcolor{blue}{hand of a bird eating food}} \\ \hline 
BART-base-P2T & {\textcolor{violet}{A bird is eating food with its hand.}} \\ \hline 
BART-base-MI & {\textcolor{orange}{The food is in the hands of a bird eating it.}} \\ \hline
Human & {\textcolor{teal}{A small bird eats food from someone's hand.}} \\
\Xhline{3\arrayrulewidth}
Concept Set & \{front, dance, routine, perform\} \\\hline
BART-base-BL & \textcolor{red}{A woman performs a routine in front of a dance.} \\ \hline
BART-base-KW & {\textcolor{brown}{A man performs a routine in front of a group of people.}} \\ \hline 
BART-base-Att & {\textcolor{blue}{A man is performing a routine in front of a group of people.}} \\ \hline 
BART-base-P2T & {\textcolor{violet}{A woman performs a routine in front of a group of people.}} \\ \hline 
BART-base-MI & {\textcolor{orange}{In this dance, a man performs a routine in front of a mirror.}} \\ \hline
Human & {\textcolor{teal}{The girl performed her dance routine in front of the audience.}} \\
\Xhline{3\arrayrulewidth}
Concept Set & \{chase, ball, owner, dog, throw\} \\\hline
BART-base-BL & \textcolor{red}{A dog is throwing a ball into a chase.} \\ \hline
BART-base-KW & {\textcolor{brown}{A dog is about to throw a ball to its owner.}} \\ \hline 
BART-base-Att & {\textcolor{blue}{A dog is trying to throw a ball at its owner.}} \\ \hline 
BART-base-P2T & {\textcolor{violet}{A dog is chasing the owner of a ball.}} \\ \hline 
BART-base-MI & {\textcolor{orange}{The dog was trained to throw balls and the dog would chase after the owner.}}\\ \hline
Human & {\textcolor{teal}{The owner threw the ball for the dog to chase after.}} \\
\Xhline{3\arrayrulewidth}
Concept Set & \{music, dance, room, listen\} \\\hline
BART-large-BL & \textcolor{red}{A listening music and dancing in a dark room} \\ \hline
BART-large-KW & {\textcolor{brown}{A group of people dance and listen to music in a room.}} \\ \hline 
BART-large-Att & {\textcolor{blue}{A group of people are dancing and listening to music in a room.}} \\ \hline 
BART-large-P2T & {\textcolor{violet}{Two people are dancing and listening to music in a dark room.}} \\ \hline 
BART-large-MI & {\textcolor{orange}{Music and dancing in the dance floor.}} \\ \hline
Human & {\textcolor{teal}{A boy danced around the room while listening to music.}}\\
\Xhline{3\arrayrulewidth}
Concept Set & \{cheer, team, crowd, goal\} \\\hline
T5-base-BL & \textcolor{red}{the crowd cheered after the goal.} \\ \hline
T5-base-KW & {\textcolor{brown}{the crowd cheered after the goal by football team}} \\ \hline 
T5-base-Att & {\textcolor{blue}{the crowd cheered after the goal by the team.}} \\ \hline 
T5-base-P2T & {\textcolor{violet}{the crowd cheered as the team scored their first goal.}} \\ \hline 
BART-base-MI & {\textcolor{orange}{The team and the crowd cheered after the goal.}} \\ \hline
Human & {\textcolor{teal}{The crowd cheered when their team scored a goal.}} \\
\Xhline{3\arrayrulewidth}
Concept Set & \{bag, put, apple, tree, pick\} \\\hline
T5-base-BL & \textcolor{red}{A man puts a bag of apples on a tree.} \\ \hline
T5-base-KW & {\textcolor{brown}{A man puts a bag under a tree and picks an apple.}} \\ \hline 
T5-base-Att & {\textcolor{blue}{A man puts a bag under a tree and picks an apple.}} \\ \hline 
T5-base-P2T & {\textcolor{violet}{A man puts a bag of apples on a tree and picks them.}} \\ \hline 
BART-base-MI & {\textcolor{orange}{A man puts a bag of apple juice on a tree to pick it up}} \\ \hline
Human & {\textcolor{teal}{I picked an apple from the tree and put it in my bag.}} \\
\Xhline{3\arrayrulewidth}
Concept Set & \{circle, ball, throw, turn, hold\} \\\hline
T5-large-BL & \textcolor{red}{Someone turns and throws a ball in a circle.} \\ \hline
T5-large-KW & {\textcolor{brown}{A man holds a ball and turns to throw it into a circle.}} \\ \hline 
T5-large-Att & {\textcolor{blue}{A man holds a ball in a circle and throws it.}} \\ \hline 
T5-large-P2T & {\textcolor{violet}{A man holds a ball, turns and throws it into a circle.}} \\ \hline 
BART-large-MI & {\textcolor{orange}{He turns and throws a ball into the circle to hold it.}} \\ \hline
Human & {\textcolor{teal}{A girl holds the ball tightly, then turns to the left and throws the ball into the net which is in the shape of a circle.}} \\
\Xhline{3\arrayrulewidth}
Concept Set & \{hair, sink, lay, wash\} \\\hline
T5-large-BL & \textcolor{red}{A woman is washing her hair in a sink.} \\ \hline
T5-large-KW & {\textcolor{brown}{A woman lays down to wash her hair in a sink.}} \\ \hline 
T5-large-Att & {\textcolor{blue}{A man lays down to wash his hair in a sink.}} \\ \hline 
T5-large-P2T & {\textcolor{violet}{A woman is washing her hair in a sink.}} \\ \hline 
BART-large-MI & {\textcolor{orange}{A woman is washing her hair in the sink. She lay the sink down}}\\\hline
Human & {\textcolor{teal}{The woman laid back in the salon chair, letting the hairdresser wash her hair in the sink.}} \\
\Xhline{3\arrayrulewidth}
Concept Set & \{wash, dry, towel, face\} \\\hline
T5-large-BL & \textcolor{red}{A man is washing his face with a towel.} \\ \hline
T5-large-KW & {\textcolor{brown}{A man washes his face with a towel and then dries it.}} \\ \hline 
T5-large-Att & {\textcolor{blue}{A man is washing his face with a towel and drying it.}} \\ \hline 
T5-large-P2T & {\textcolor{violet}{A man is washing his face with a towel and drying it off.}} \\ \hline 
BART-large-MI & {\textcolor{orange}{A man is washing his face with a towel to dry it.}} \\ \hline
Human & {\textcolor{teal}{The woman will wash the baby's face and dry it with a towel.}} \\
\Xhline{3\arrayrulewidth}
\end{tabular}}
\caption{\footnotesize Qualitative examples for test$_{CG}$. Color coded: \textcolor{red}{baseline}, \textcolor{brown}{Kw-aug}, \textcolor{blue}{Att-aug}, \textcolor{violet}{P2T}, \textcolor{orange}{MI}, and \textcolor{teal}{human reference}.}
\label{tab:qualitative_appendix}
\vspace{-2mm}
\end{table}

\begin{table*}[h]
\centering
\small
\begin{tabular}{|c|cc|cc|c|cc|c|c|}
\hline
 \textbf{Model\textbackslash Metrics} & \multicolumn{2}{c|}{ROUGE-2/L} & \multicolumn{2}{c|}{BLEU-3/4} & METEOR & CIDEr & SPICE & BERTScore & Cov\\
 \hline
 Reported BART-large & 22.13 & 43.02 & 37.00 & 27.50 & 31.00 & 14.12 & 30.00 & - & 97.56 \\
 \hline
 Reported T5-base & 15.33 & 36.20 & 28.10 & 18.00 & 24.60 & 9.73 & 23.40 & - & 83.77 \\
 \hline
 Reported T5-Large & 21.98 & 44.41 & 40.80 & 30.60 & 31.00 & 15.84 & 31.80 & - & 97.04 \\
 \thickhline
 Our BART-base & 15.91 & 36.15 & 38.30 & 28.30 & 30.20 & 15.07 & 30.35 & 58.26 & 93.44 \\
 \hline
 Our BART-large & 17.27 & 37.32 & \textbf{39.95} & \textbf{30.20} & \textbf{31.15} & \textbf{15.72} & \textbf{31.20} & 58.58 & 95.03 \\ 
 \hline
 Our T5-base & \textbf{17.27} & \textbf{37.69} & \textbf{41.15} & \textbf{31.00} & \textbf{31.10} & \textbf{16.37} & \textbf{32.05} & 60.32 & \textbf{94.44} \\
 \hline
 Our T5-large & 17.90 & 38.31 & \textbf{43.80} & \textbf{33.60} & \textbf{32.70} & \textbf{17.02} & \textbf{33.45} & 61.39 & 96.26 \\
 \hline
\end{tabular}
\caption{\footnotesize Performance of our re-implemented CommonGen models on dev$_{O}$ compared to the original numbers reported in \citet{lin-etal-2020-commongen}. Note that for our models, results are averaged over two seeds, and that the original authors did not experiment with BART-base. Bold indicates where we match or exceed the reported metric.}
\label{tab:full_reimplementation_stats}
\end{table*}



\section{Human Evaluation Details}\label{sec:appendix_human_eval}
\begin{figure*}[!ht]
\begin{subfigure}{.97\textwidth}
    \centering
    \includegraphics[width=0.99\textwidth]{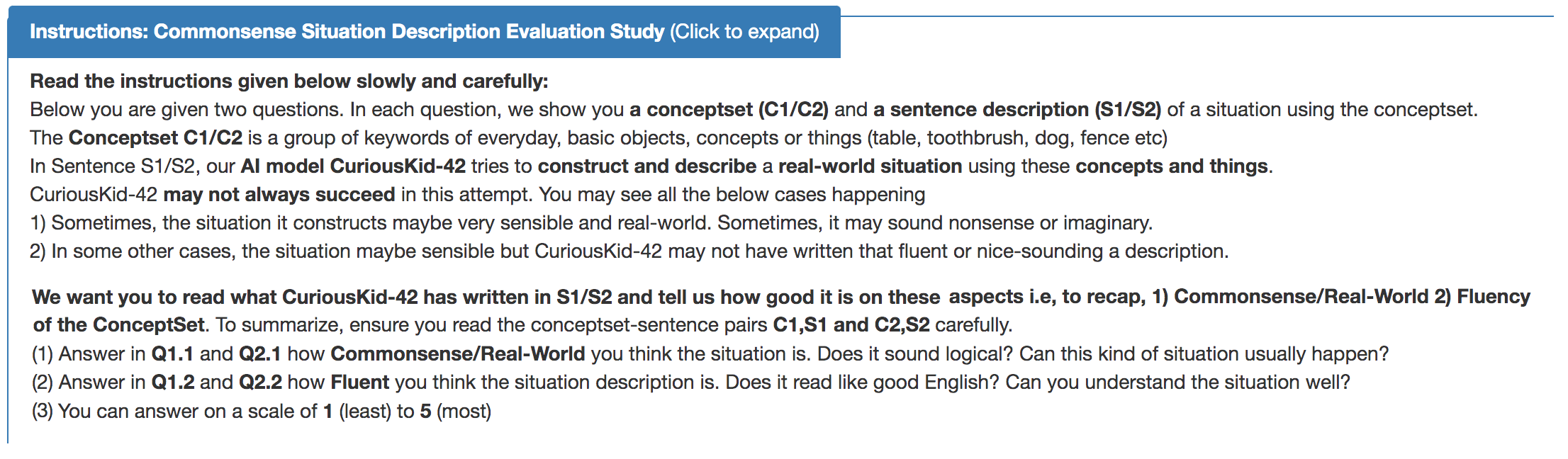}
    \caption{}
    \label{fig:amt_preserve_instructions_graph}
\end{subfigure} \\
\begin{subfigure}{.97\textwidth}
    \centering
    \includegraphics[width=0.99\textwidth]{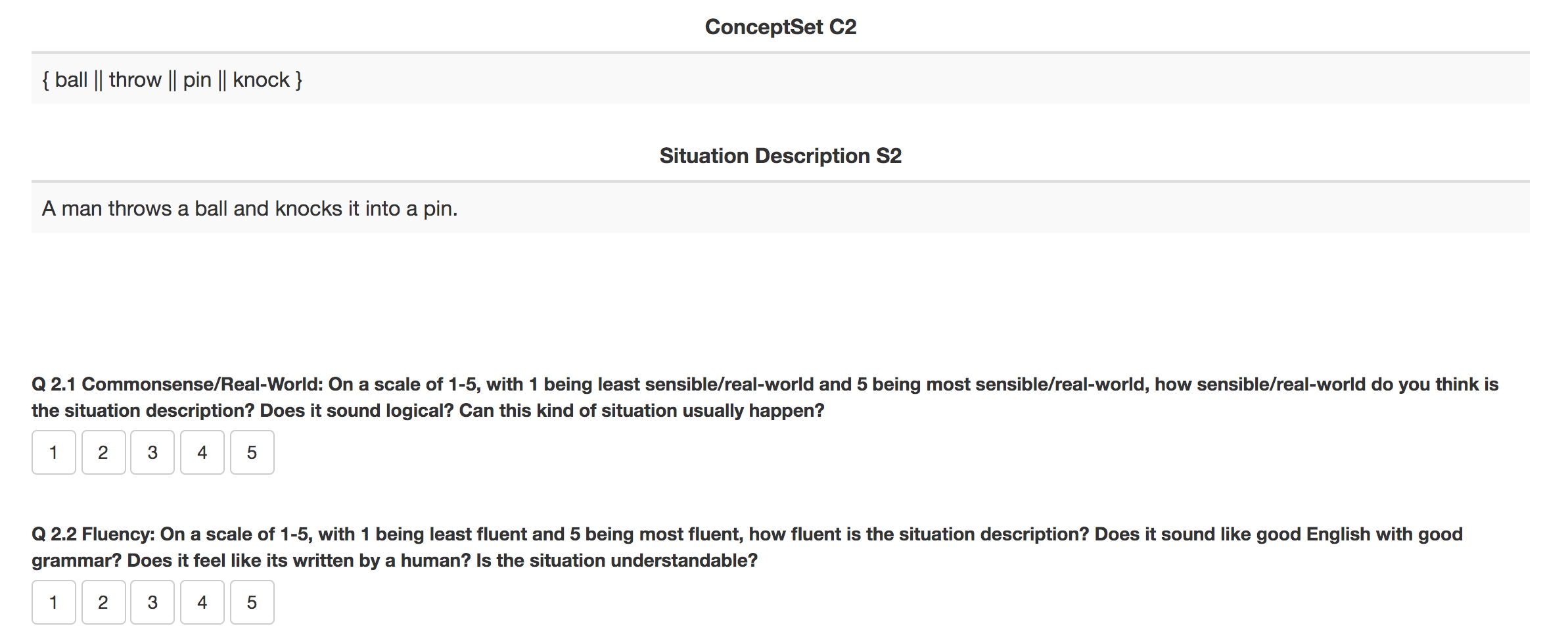}
    \caption{}
    \label{fig:amt_preserve_questions_graph}
\end{subfigure}
\vspace{-0.8\abovedisplayskip}
\vspace{1mm}
\caption{\small Snapshots of human evaluation: a) instructions seen by annotator and b) an example with questions.\label{fig:human_eval_template}}
\end{figure*}

Human evaluation was done via paid crowdworkers on AMT, who were from Anglophone countries with lifetime approval rates $>97$\% . Each example was evaluated by 2 annotators. The time given for each AMT task instance or HIT was 8 minutes. Sufficient time to read instructions, as calibrated by authors, was also considered in the maximum time limit for performing each HIT. Annotators were paid 98 cents per HIT. This rate (7.35\$/hr) exceeds the minimum wage for the USA (7.2\$/hr) and constitutes fair pay. We neither solicit, record, request, or predict any personal information pertaining to the AMT crowdworkers. Specific instructions and a question snippet can be seen in Figure \ref{fig:human_eval_template}.

\section{Graphs Displaying Other Hyperparameter Results}\label{appendix:hyperparam_graphs}
\begin{figure*}[t]
    \centering
    \includegraphics[width=0.32\textwidth]{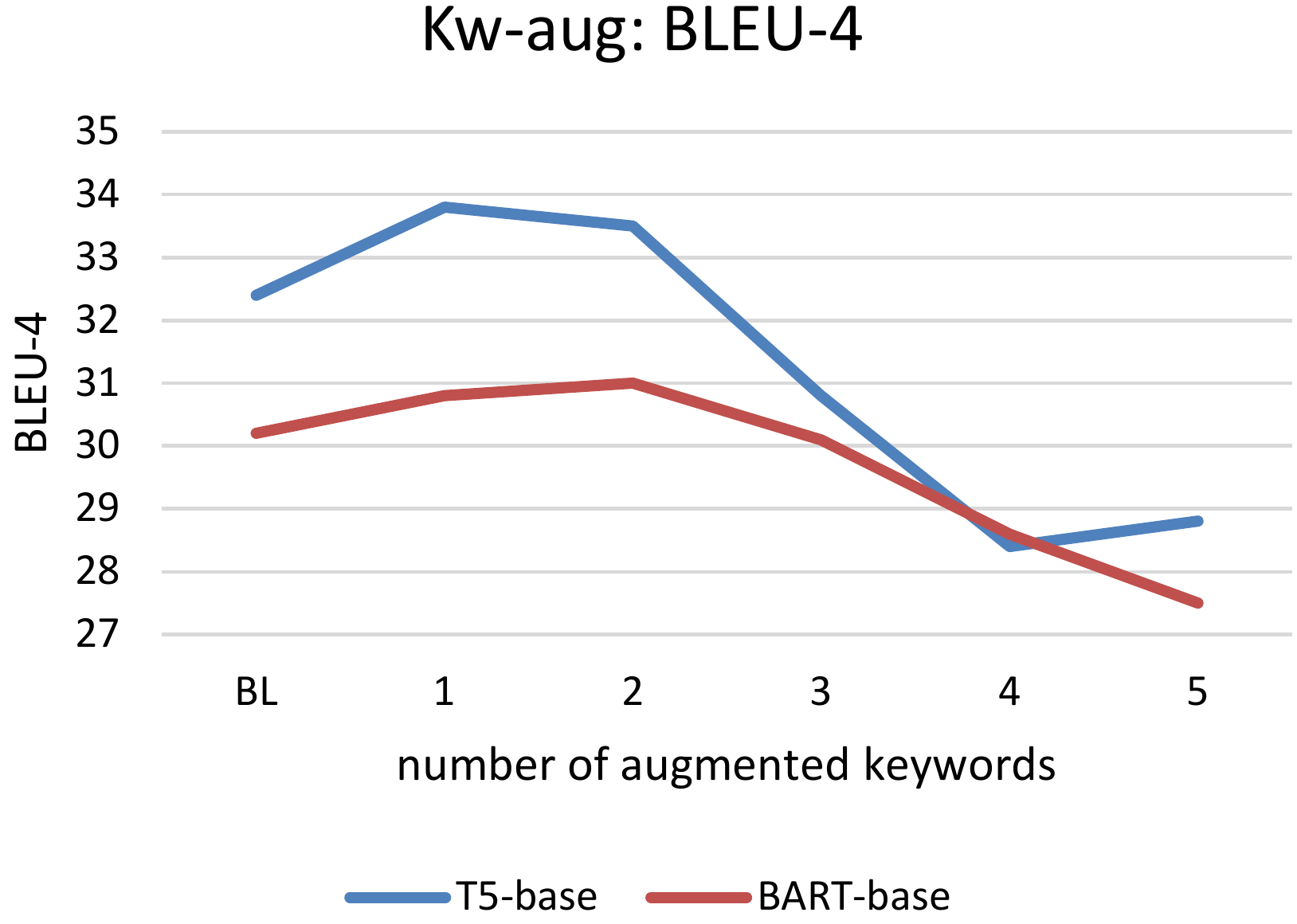}
    \includegraphics[width=0.32\textwidth]{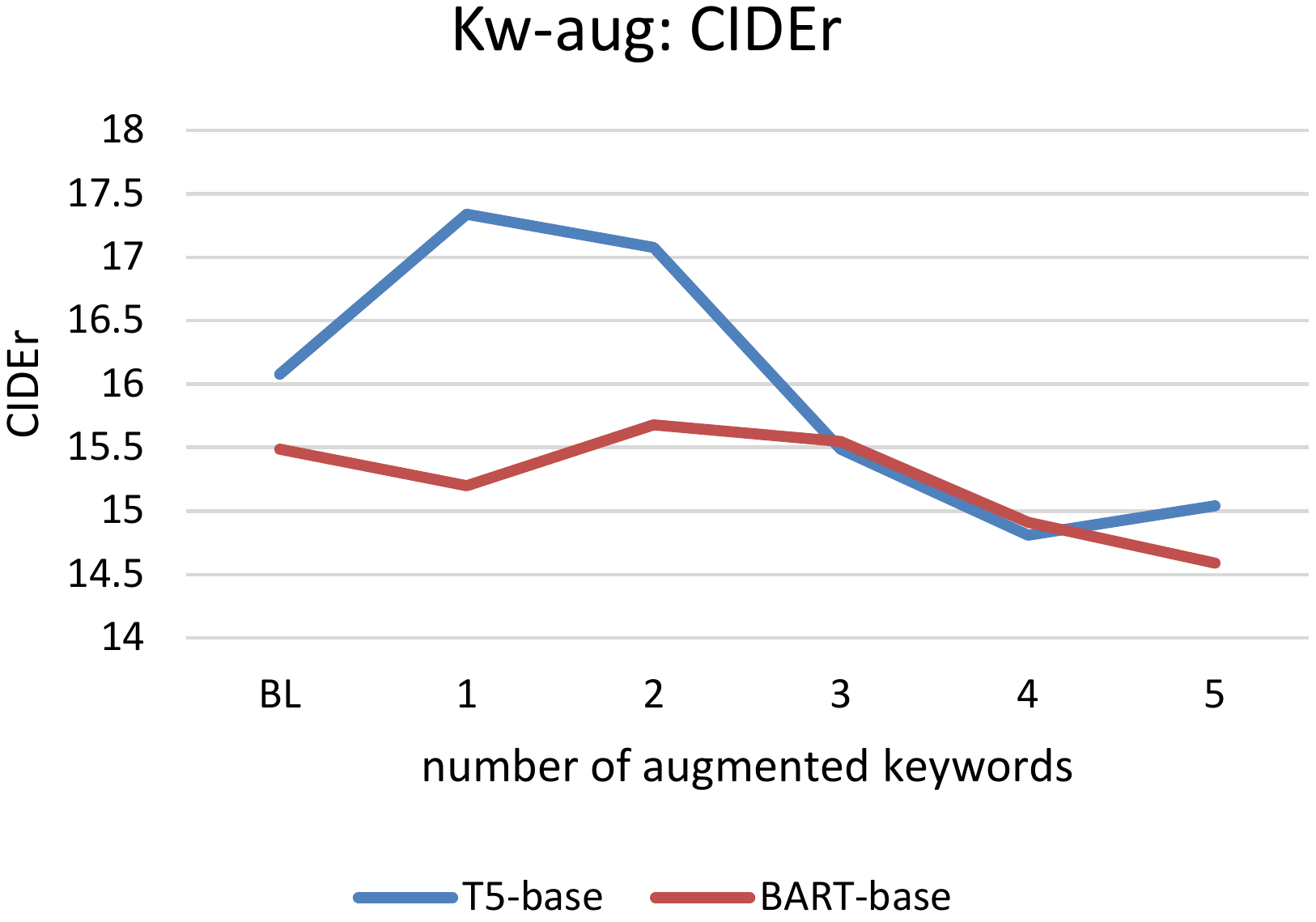}
    \includegraphics[width=0.32\textwidth]{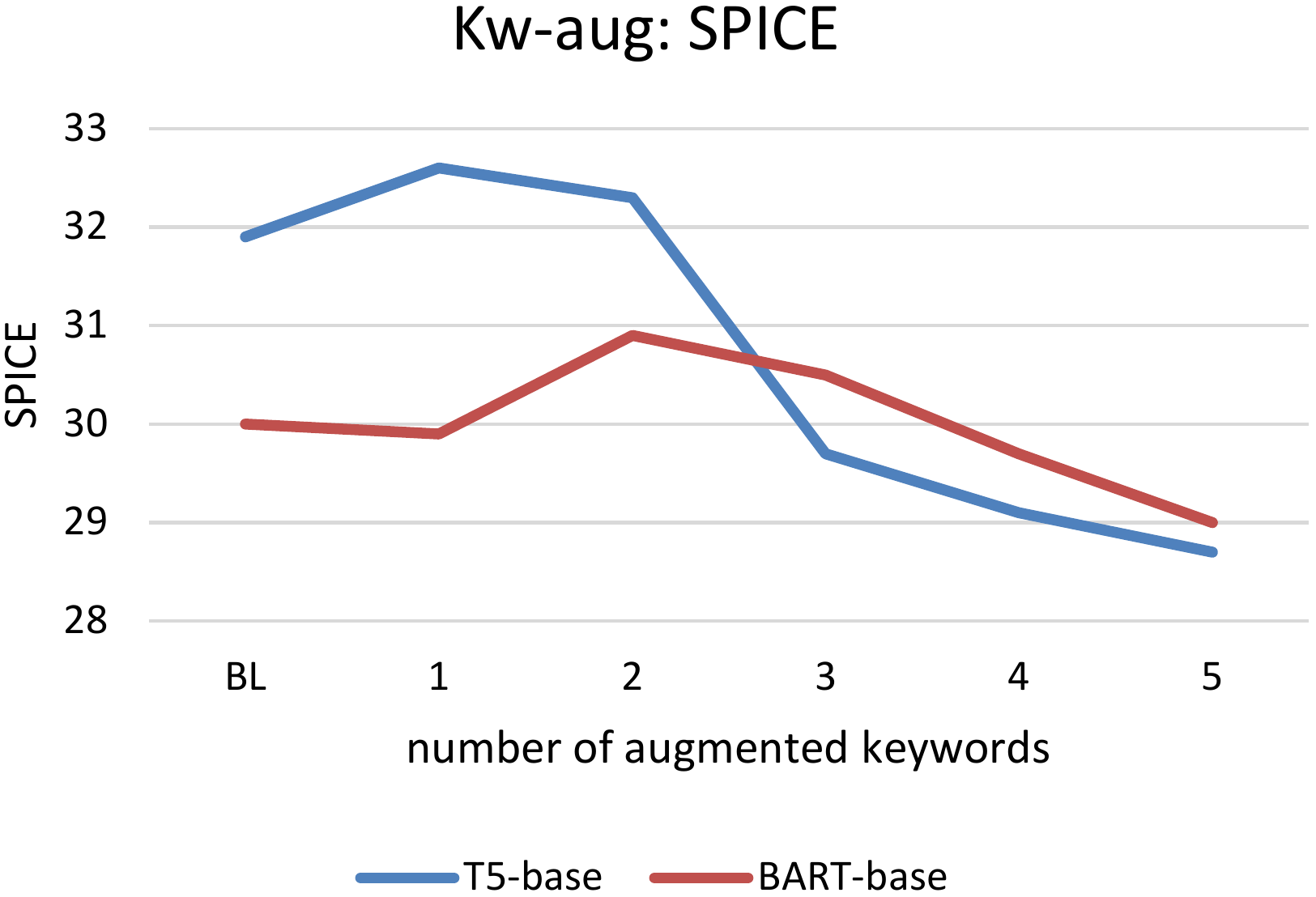}
    \caption{\footnotesize Kw-aug: graphs of BLEU-4, CIDEr, and SPICE results on test$_{CG}$ over different numbers of augmented keywords for BART-base and T5-base. These are only first seed results, and we only went above three augmented keywords for the base size models. BL refers to the baseline results with no augmented keywords.} \label{fig:Kw-aug_graphs}
\vspace{-3mm}
\end{figure*}

\begin{figure*}[t]
    \centering
    \includegraphics[width=0.32\textwidth]{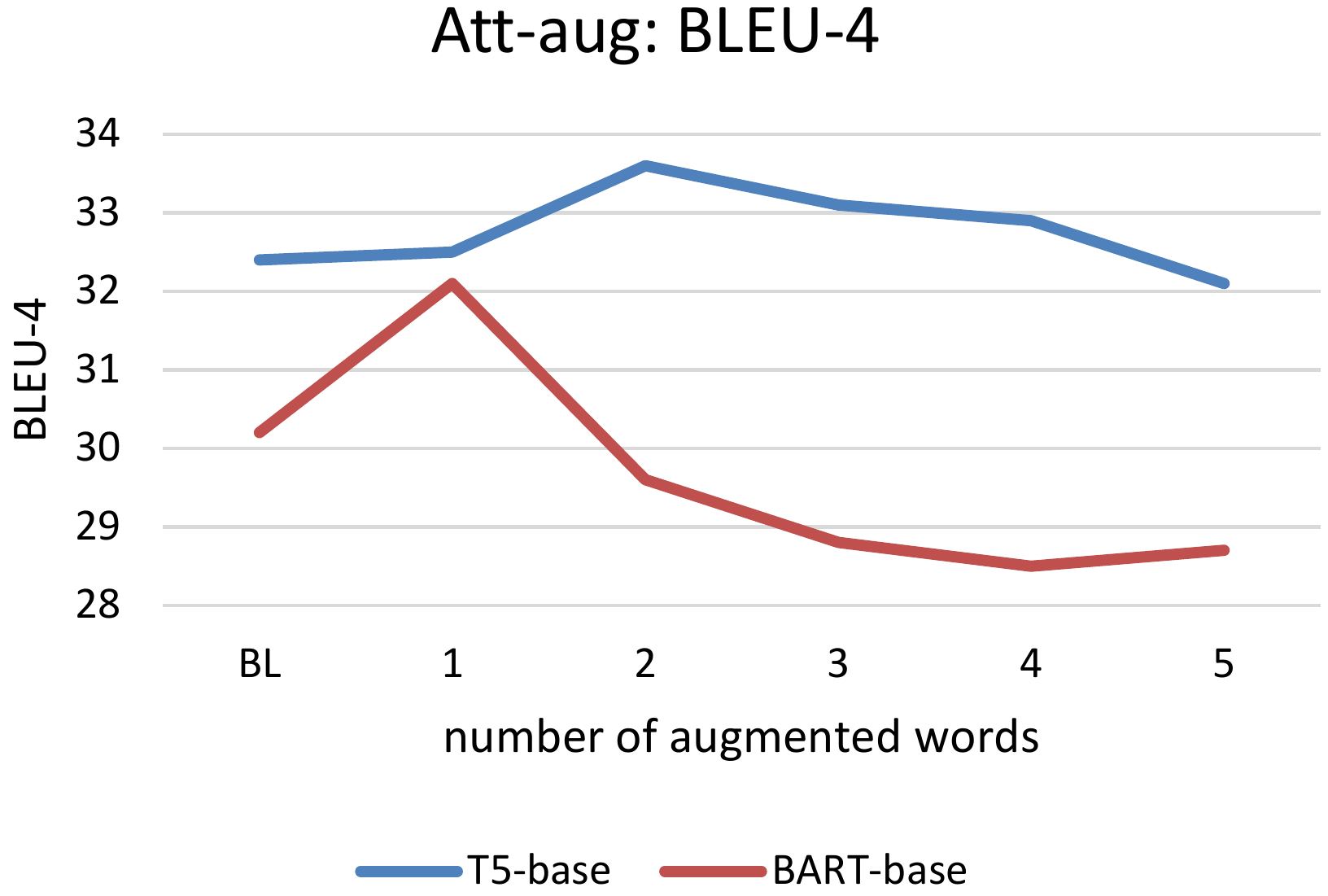}
    \includegraphics[width=0.32\textwidth]{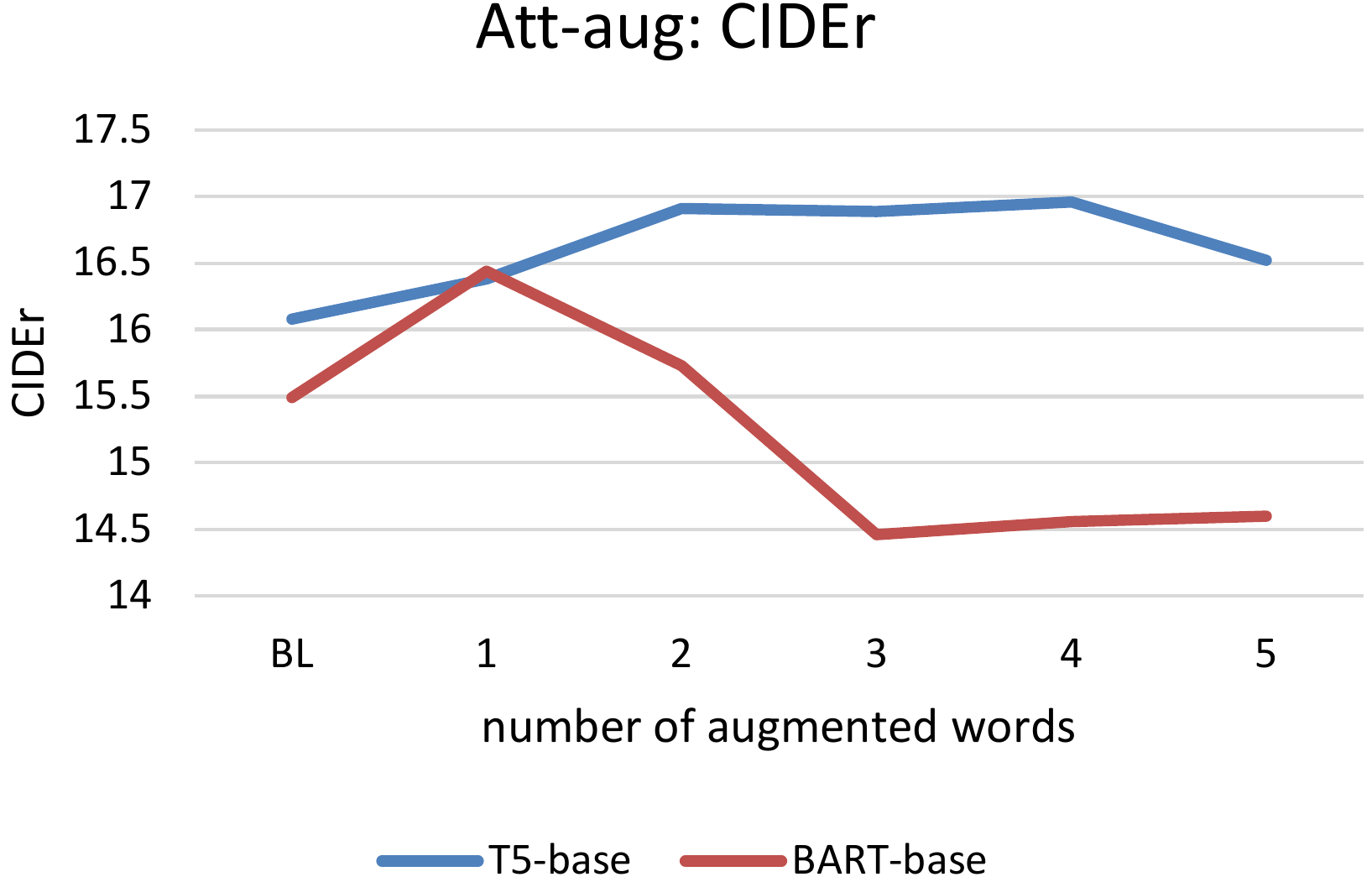}
    \includegraphics[width=0.32\textwidth]{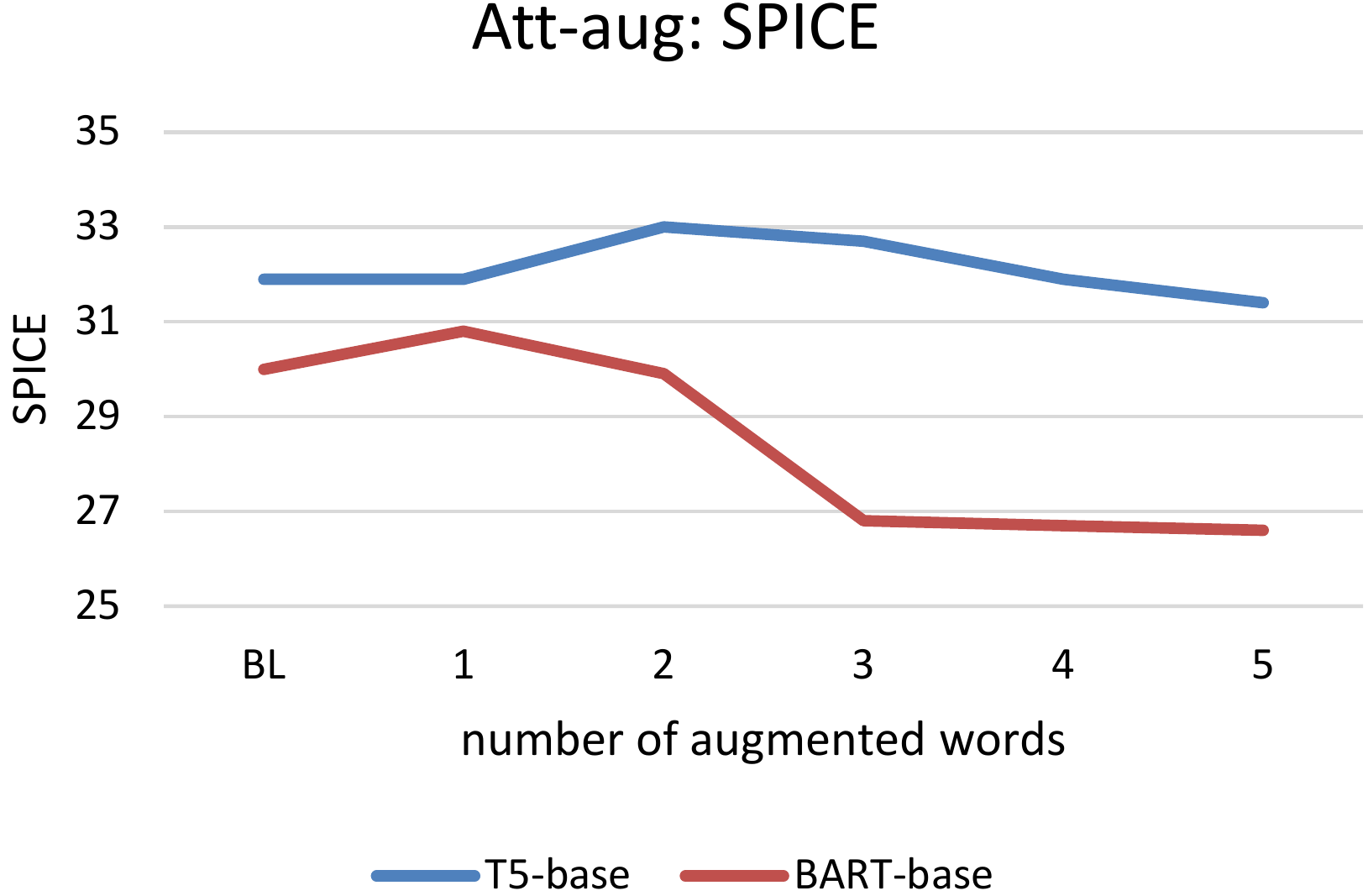}
    \caption{\footnotesize Att-aug: graphs of BLEU-4, CIDEr, and SPICE results on test$_{CG}$ over different numbers of augmented words for BART-base and T5-base. These are only first seed results, and we only went above three augmented words for the base size models. BL refers to the baseline results with no augmented words.} \label{fig:Att-aug_graphs}
\vspace{-3mm}
\end{figure*}

\begin{figure*}[t]
    \centering
    \includegraphics[width=0.32\textwidth]{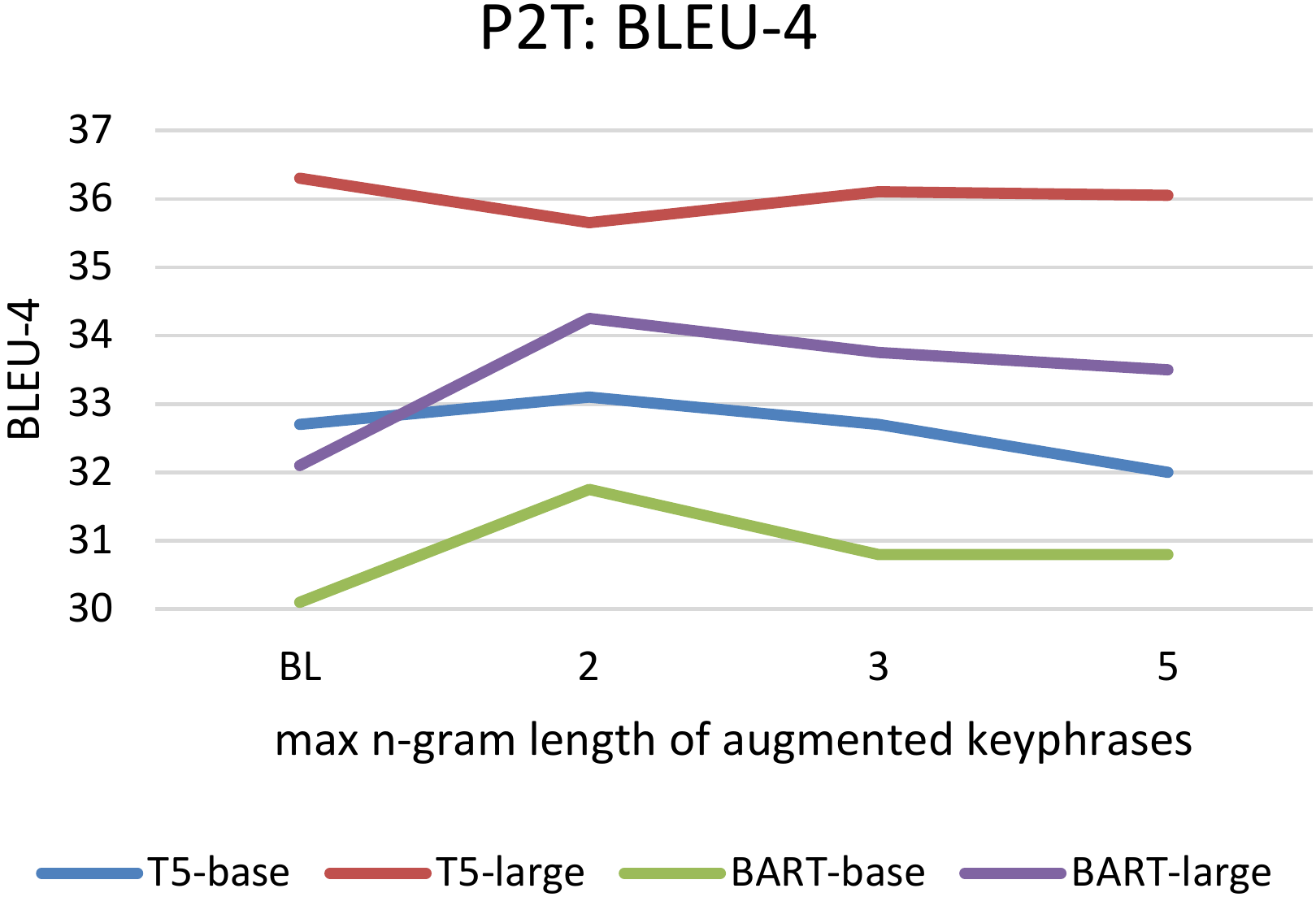}
    \includegraphics[width=0.32\textwidth]{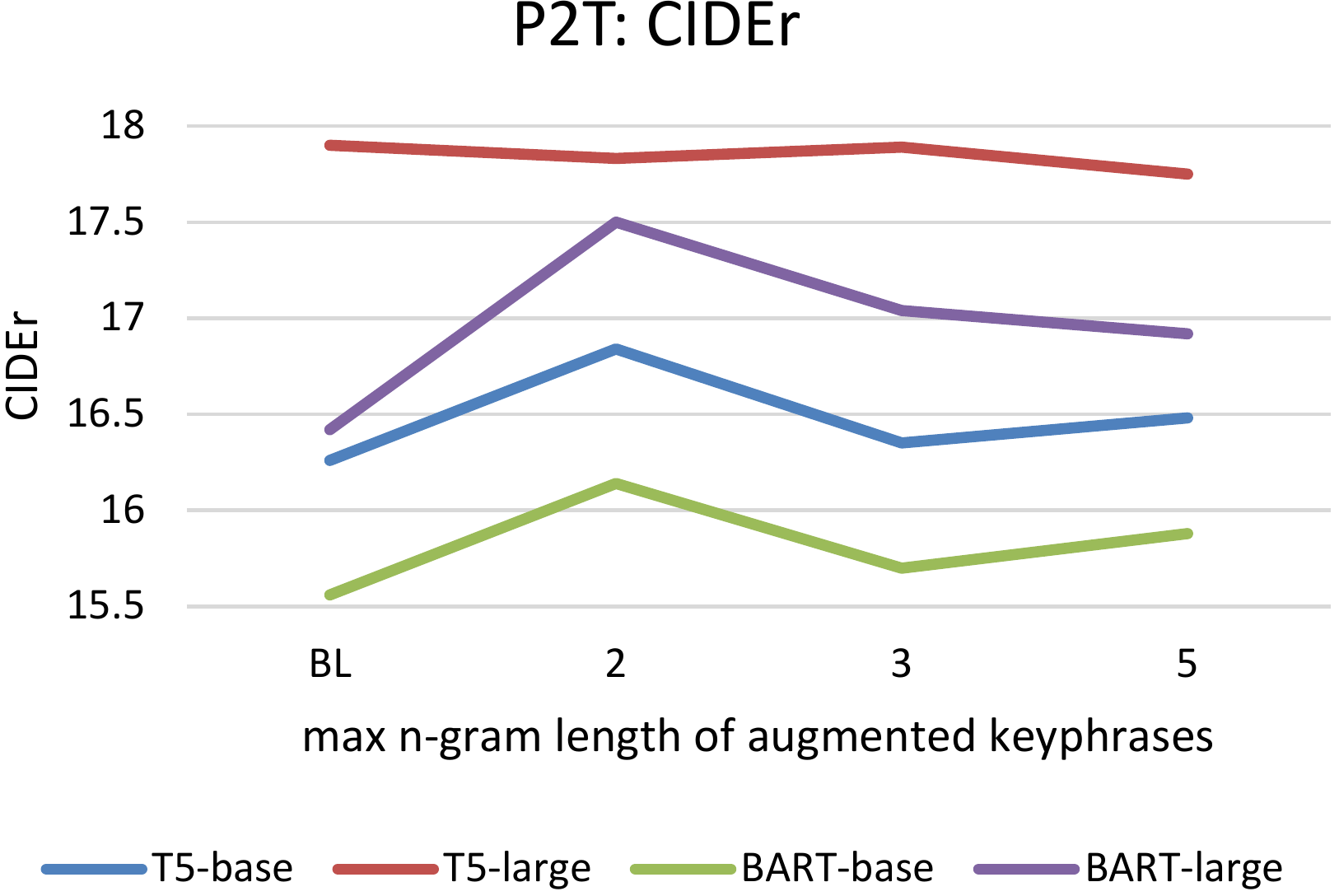}
    \includegraphics[width=0.32\textwidth]{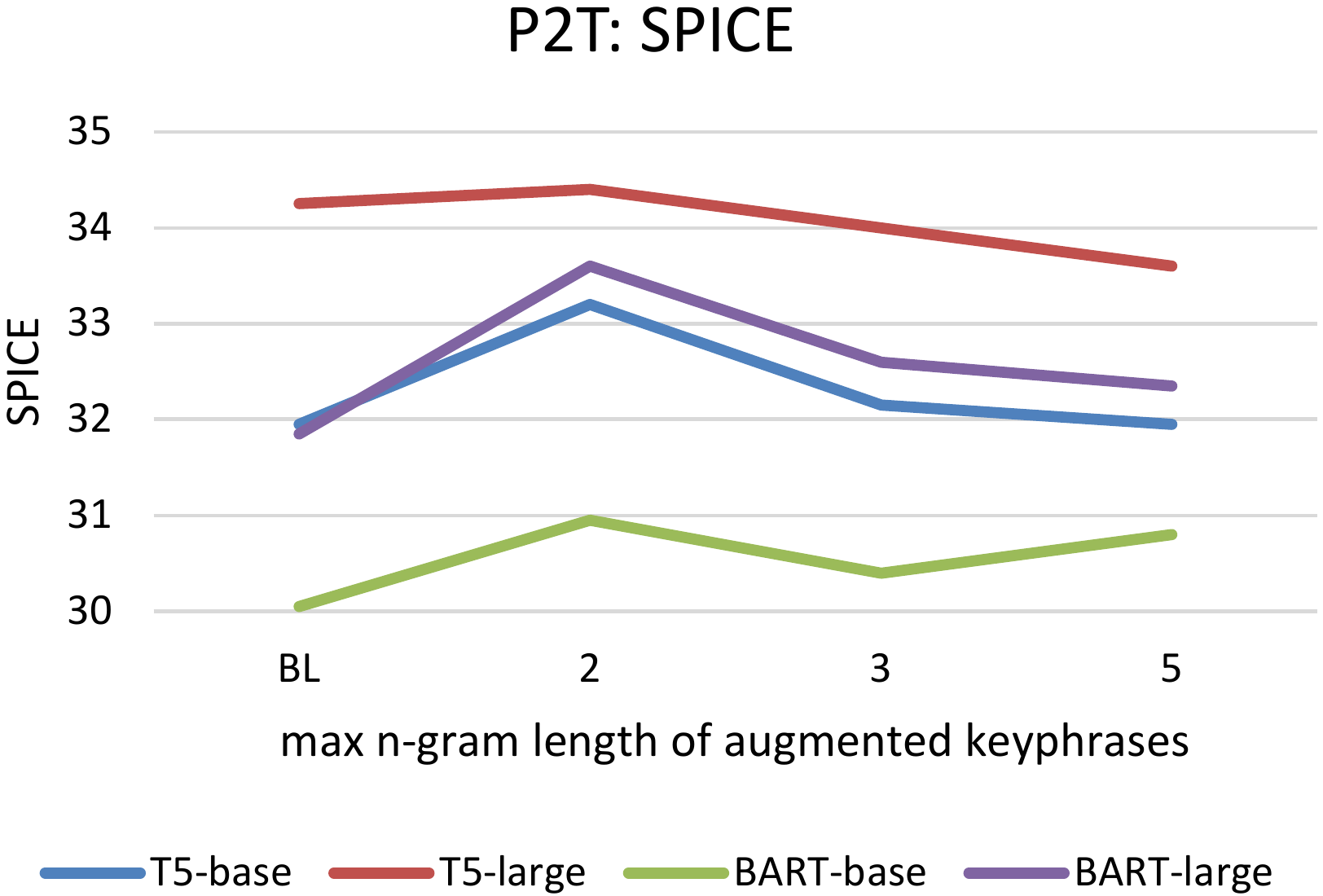}
    \caption{\footnotesize P2T: graphs of BLEU-4, CIDEr, and SPICE results on test$_{CG}$ over different max n-gram lengths of augmented keyphrases. These are results averaged over two seeds. BL refers to the baseline results with no augmented keyphrases.} \label{fig:P2T_graphs}
\vspace{-3mm}
\end{figure*}

\begin{figure*}[t]
    \centering
    \includegraphics[width=0.32\textwidth]{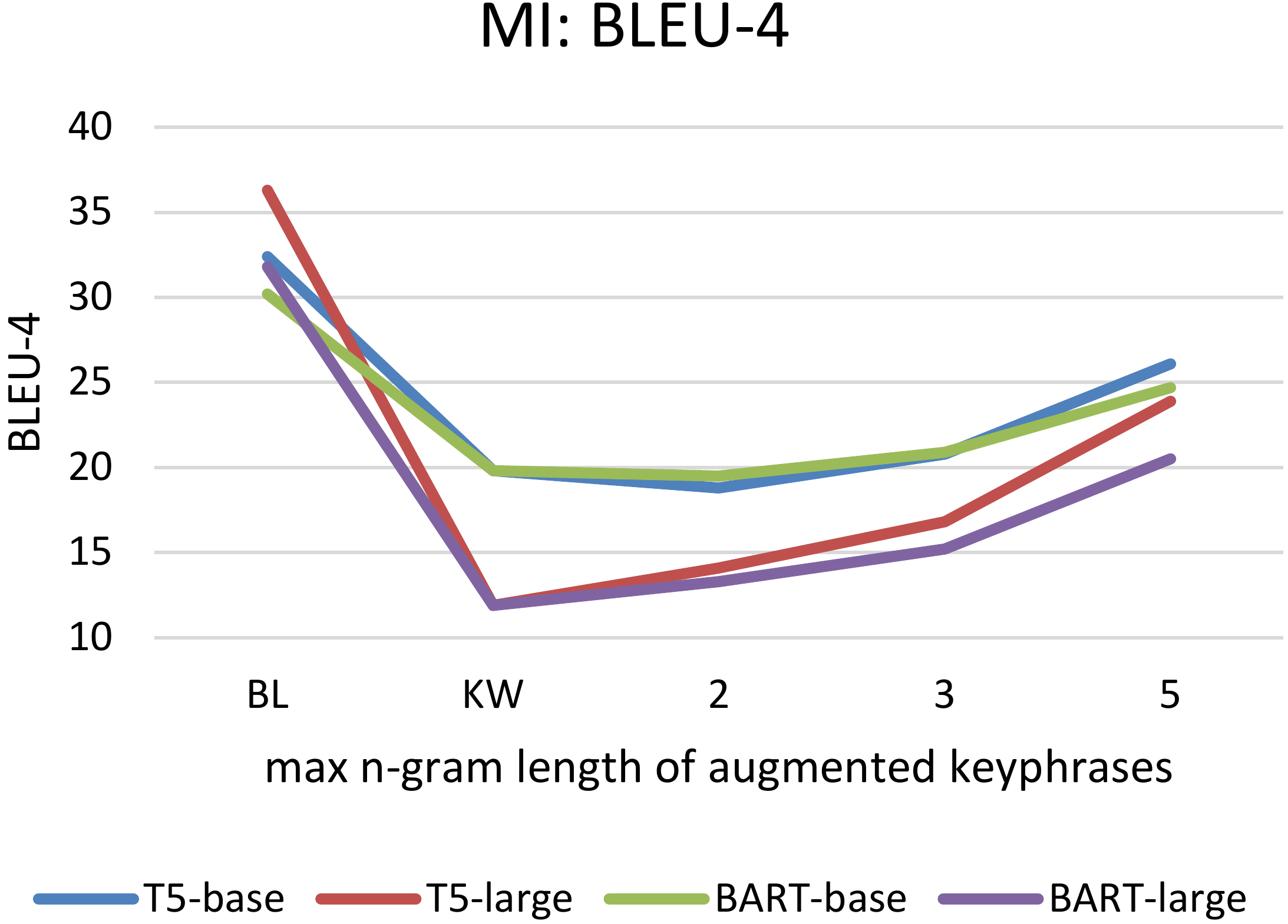}
    \includegraphics[width=0.32\textwidth]{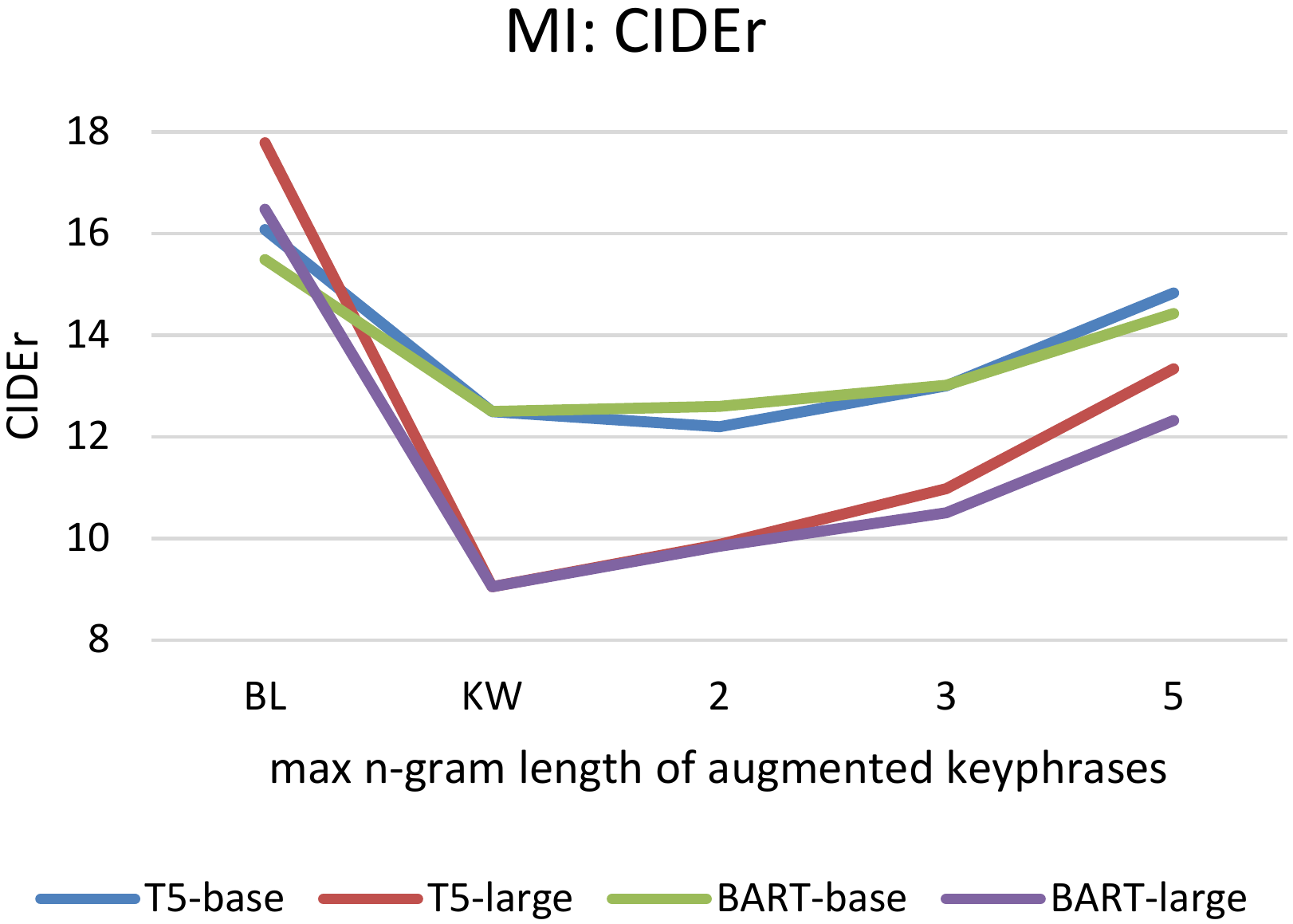}
    \includegraphics[width=0.32\textwidth]{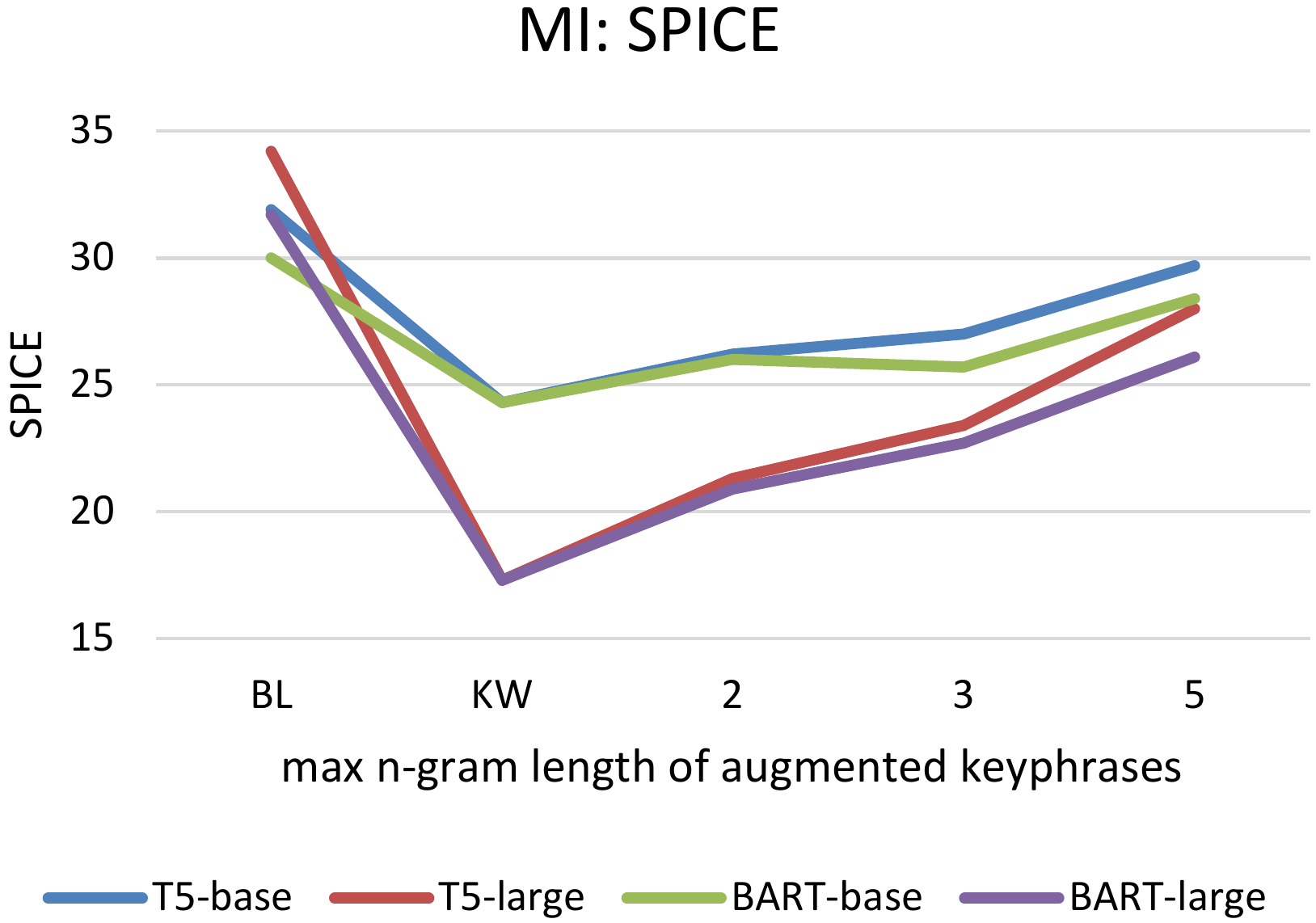}
    \caption{\footnotesize Mask Infilling (MI): graphs of BLEU-4, CIDEr, and SPICE results on test$_{CG}$ over different max n-gram lengths of augmented keyphrases. These are first seed results only. BL refers to the baseline results, and KW refers to mask infilling on the original keywords only (with no augmented keyphrases).} \label{fig:MI_graphs}
\end{figure*}

Figures \ref{fig:Kw-aug_graphs}, \ref{fig:Att-aug_graphs}, \ref{fig:P2T_graphs}, and \ref{fig:MI_graphs} contain graphs displaying other hyperparameter results for Kw-aug, Att-aug, P2T, and Mask Infilling (MI), respectively.

\section{Further Qualitative Examples}\label{sec:appendix_qualitative_examples}
See Table \ref{tab:qualitative_appendix} for further qualitative examples.

\end{document}